\documentclass{ametsocV6.1}
\usepackage{booktabs}
\usepackage{lineno}
\usepackage{subcaption}
\usepackage{tikz}
\usetikzlibrary{positioning, arrows.meta, shapes.geometric}
\usepackage{multirow}

\let\internallinenumbers\relax

\title{Comparing skill of historical 
rainfall data based monsoon rainfall prediction in India
with NWP forecasts}

\makeatletter
\let\@LN@ppage\relax
\let\do@mlinenumbers\relax
\makeatother

\begin{document}

\authors{Aastha Jain,\aff{a}\correspondingauthor{Aastha Jain, aasthajn25@gmail.com} 
Apoorva Narula,\aff{b} 
Jatin Batra,\aff{c} 
M Rajeevan,\aff{d} 
Sandeep Juneja \aff{a} 
}

\affiliation{\aff{a}{Ashoka University}\\
\aff{b}{Indian Institute of Technology Delhi}\\
\aff{c}{Tata Institute of Fundamental Research}\\
\aff{d}{Atria University}
}

\abstract{
The Indian summer monsoon is a highly complex and critical weather system that directly affects the livelihoods of over a billion people across the Indian subcontinent. Accurate short-term forecasting remains a major scientific challenge due to the monsoon's intrinsic nonlinearity and its sensitivity to multi-scale drivers, including local land-atmosphere interactions and large-scale ocean-atmosphere phenomena. In this study, we address the problem of forecasting daily rainfall across India during the summer months, focusing on both one-day and three-day lead times. We use Autoformers - deep learning transformer-based architectures designed for time series forecasting. These are trained on historical gridded precipitation data from the Indian Meteorological Department (1901--2023) at spatial resolutions of $0.25^\circ \times 0.25^\circ$, as well as $1^\circ \times 1^\circ$. The models also incorporate auxiliary meteorological variables from ECMWF’s reanalysis datasets, namely, cloud cover, humidity, temperature, soil moisture, vorticity, and wind speed.
Forecasts at $0.25^\circ \times 0.25^\circ$ are benchmarked against ECMWF’s High-Resolution Ensemble System (HRES), widely regarded as the most accurate numerical weather predictor, and at $1^\circ \times 1^\circ $ with those from National Centre for Environmental Prediction (NCEP). We conduct both nationwide evaluations and localized analyses for major Indian cities. Our results indicate that transformer-based deep learning models consistently outperform both HRES and NCEP,  as well as other climatological baselines. Specifically, compared to our model, forecasts from HRES and NCEP model have about 22\% and 43\% higher error, respectively,
for a single day prediction, and over 27\% and 66\% higher error respectively, for a three day prediction.  Persistence-based predictions show a 40\% and 69\% higher error for one-day and three-day forecasts, respectively.
Moreover, our models demonstrate superior skill in balancing heavy rainfall detection with false positives. We also find that incorporating historical data up to 20 days prior significantly reduces forecast error, particularly in landlocked regions.
Our findings suggest that NWP forecasts for the Indian monsoon can be substantially improved by integrating diverse, high-resolution observational data with carefully designed deep learning models tailored to monsoon dynamics.
}

\maketitle

%
%
%

%
%

%
\section{Introduction}
Accurate rainfall prediction in India during monsoons is crucial for a variety of reasons: agriculture planning, disaster management, day-to-day transportation planning, and so on. Anecdotally, it is well known that numerical weather prediction (NWP) does not perform well in the prediction of rainfall for India \citep{Hindu_article}. It is also conjectured that during monsoons, rainfall data across India have spatial-temporal memory so that information on rainfall early in neighboring parts may be useful for future rainfall prediction \citep{goswami2003breaks}. In addition, rainfall has been shown to be also affected by a variety of other atmospheric, land, and ocean variables, such as temperature, wind, soil moisture, etc. \citep{prasad1988large}.\\
In this paper, we consider daily gridded precipitation data from India Meteorological Department (IMD) \citep{pai2014development} available from 1901--2023, at a spatial resolution of $0.25^\circ \times 0.25^\circ$, as well as $1^\circ \times 1^\circ$ (each degree roughly corresponds to 111 km).
We use this to predict rainfall for all of India, one day and three days in the future. We also use daily atmospheric and land data as additional covariates in an attempt to improve our forecasts. We compare our performance with operational NWP forecasts including HRES-IFS (High Resolution Integrated Forecast System) from ECMWF (European Centre for Medium-Range Weather Forecasts) \citep{https://doi.org/10.21957/open-data}, and those from National Centre for Environmental Prediction (NCEP), \citep{https://doi.org/10.24381/cds.181d637e}. HRES is widely regarded as the top operational weather forecasting system in the world \citep{graphcast}. NCEP is used as it forms the base model for IMD Global Forecasting System ( \cite{IMD_SOP})\\
Several attempts have been made to predict rainfall using machine learning (ML) techniques. For long-range forecasting of monsoon rainfall in India, \cite{rajeevan2000new, rajeevan2007new} used a host of methods such as multivariate principal component regression, simple neural networks, linear discriminant analysis, ensemble multiple linear regression, and projection pursuit regression. They used multimodal data such as air temperature, sea surface temperature, rainfall, air pressure etc. These developments helped support IMD’s two-stage monsoon forecasting system with the first stage forecast given in mid April and an updated second stage forecast given at the end of June.\\
More recently, deep learning and machine learning approaches have been explored for short-range rainfall prediction. \cite{kumar2021deep, kumar2022deep} conducted a comparative analysis of Long-Short-Term-Memory (LSTM) and ConvLSTM models trained using ground-based IMD rainfall data and satellite data for Indian summer monsoon rainfall. They showed correlation coefficients between the observed and predicted rainfall of 0.67 for 1 day and 0.42 for 2 day lead time, respectively, indicating reasonable skill in short-range precipitation forecasting. However, their results also highlighted that model efficiency quickly drops after 2 days lead time, pointing to a limitation in capturing longer temporal dependencies. Similarly, \cite{praveen2020analyzing} analyzed rainfall trends and forecasting using machine learning. They  mainly focus on predicting long-term trends. \cite{jose2022improving} developed ensemble predictions of daily precipitation and temperature using machine learning. Their analysis was limited to coarser spatial scales ($1^\circ$ resolution). \cite{miao2019improving} used a combined CNN-LSTM neural network to improve the prediction of monsoon precipitation. The last three references relied on reanalysis data for ground truth. 
Typically, both the NWP based prediction models and machine learning based prediction models, such as GraphCast \citep{graphcast} and ClimaX \citep{nguyen2023climax}, also rely on reanalysis data for initialization and, in case of ML models, training. 
In our work, we instead use the IMD data for training, which is shown to be better representative of the ground truth \citep{kishore2016precipitation}.\\
 \cite{chen2023machine} provide a comprehensive survey of machine learning methods in weather and climate applications and highlight persistent challenges such as the underprediction of extreme rainfall events and the need for better integration with physical models.\\

Our work adds to the growing literature on the use of ML for short-term rainfall prediction, primarily using historical IMD data and benchmarking against NWP forecasts. We compare and contrast the deep learning-based forecasts generated by autoformers \citep{wu2021autoformer} using historical rainfall data from IMD (referred to as DL-HD forecasts) and the forecasts generated using IMD rainfall data and additional covariates (called DL-HD+Covariates), with the NWP forecasts. In an attempt to arrive at improved forecasts, we also combine the NWP forecasts with DL-HD+Covariates using a simple neural network, to generate ensemble forecasts. Much of the rainfall in India occurs during the four monsoon months of June, July, August and September (JJAS), and hence, to build a more useful forecasting system, we restrict our forecasts to JJAS.\\ 
We find that for forecasting precipitation one and three days into the future, forecasts obtained by DL-HD+Covariates are substantially more accurate compared to NWP forecasts as well as forecasts based on climatological baselines. We also discuss the improvement in performance of our forecasts when they are combined with NWP forecasts. We further observe that using autoformers, data up to 20 days in the past is useful in reducing errors of one and three day forecasts.


\section{Data and Experiments}
\label{sec:data}

\subsection{Data Sources}
\label{subsec: data_sources}
\begin{enumerate}
    \item \textbf{IMD Ground Truth}: We use daily gridded precipitation data obtained from IMD spanning the period from 1901 to 2023, at a spatial resolution of $0.25^{\circ} \times 0.25^{\circ}$ \citep{pai2014development}. At this resolution, the geographical extent of India is discretized into 12,422 grids.  We also use gridded data at $1^{\circ} \times 1^{\circ}$ resolution \citep{rajeevan2008analysis} during the same period, which provides data at 357 grids. This forms the ground truth dataset against which our predictions and other models are compared.
    \item \textbf{Additional weather variables}: Apart from precipitation, we also use daily atmospheric and land data at $0.25^\circ$ resolution provided by ECMWF as part of their reanalysis products \citep{https://doi.org/10.24381/cds.e2161bac}. These variables include: horizontal and vertical components of wind at 10m, temperature, soil moisture, cloud cover, vorticity at  850hPa, humidity, and divergence at 700hPa. These are the lower tropospheric pressure levels, indicative of cloud development and rainfall processes. The data is available from 1950 onwards. 
     \item \textbf{NWP forecasts}: We conduct all comparisons against two popular NWP bechmarks, HRES and NCEP
forecasts. The HRES daily forecasts are obtained from ECMWF \citep{https://doi.org/10.21957/open-data} for all years 2011 onwards, at a resolution of $0.25^\circ$, for both 1 and 3 days into the future. The NWP forecasts from NCEP are obtained from \cite{https://doi.org/10.24381/cds.181d637e}. It is available more or less on alternate days through the years $2011-2019$, capturing about 60\% of the days. Thereafter the data is available daily. The NCEP dataset provides gridded forecasts at 6 hour intervals, for three time periods: 1, 2 and 3 days into the future. The spatial resolution of these forecasts is $1^{\circ} \times 1^{\circ}$, with predictions recorded as cumulative values over the subsequent 24 hours at 6-hour intervals. Both the datasets are downloaded only for the JJAS months.

\end{enumerate}

\subsection{Dataset Preparation}
We compare the 06:00 UTC daily NWP forecasts for lead times of 1 and 3 days with the corresponding deep-learning-based forecasts. All data are aligned to 06:00 IST to ensure proper temporal consistency with the ground truth. The NWP data provide cumulative precipitation forecasts over 24-hour and 72-hour periods, which are directly comparable to our 1-day and 3-day predictions, respectively. All datasets are available at $0.25^\circ \times 0.25^\circ$ resolution without any grid mismatch, so these can used directly. However, for $1^{\circ} \times 1^{\circ}$ comparisons simple linear interpolation is needed  to bring all datasets to the same resolution. More details on this alignment process for $1^\circ$ data are given in Appendix~\ref{app:1deg}.\\
For DL-HD and DL-HD+Covariates training, the IMD dataset is partitioned into training (1901–2011) and test (2012–2023) subsets. Training samples are constructed using a time window approach, where each input consists of rainfall data from all grid points over $d$ contiguous days. The model is trained to predict the cumulative rainfall for the $(d + 1)^\mathrm{th}$ day at the same grid points.

\subsection{Experiments}
Below we outline how forecasts are generated using different models for lead times of 1 and 3 days. 
\label{section:models}
\begin{enumerate}

    \item \textbf{DL-HD:} 
    We generate forecasts for all $n$ grids across India using historical rainfall data from IMD, utilizing varying lengths of past information, spanning from 3 to 20 days ($d$). The input dimensions for the models are structured as $n \times d$, capturing the historical rainfall data for all grids over the specified timeframe. The output dimension is $n$, representing the forecasted rainfall for the subsequent day at each grid point.

This is implemented using the Autoformer architecture, an advanced variant of the Transformer model specifically designed for time series forecasting. Autoformer incorporates two core components that make it particularly effective for modeling the complex spatiotemporal dynamics of rainfall data:
\begin{enumerate}
    \item Series decomposition: This mechanism decomposes the input time series into trend and seasonal components at multiple stages within the model. The trend component captures long-term variations in rainfall, while the seasonal component isolates shorter-term, repeating fluctuations.
    \item Auto-correlation mechanism: Replacing the standard self-attention mechanism, Autoformer employs an auto-correlation approach that identifies repeating patterns over different time horizons. This helps capture periodic dependencies and reduces computational complexity from quadratic to near-linear in sequence length.
\end{enumerate}

We train the Autoformer model using data from 1901 to 2011 and generate test forecasts for the years 2012 to 2023. More details on the Autoformer architecture we use in our experiments is given in Appendix \ref{app:nn_hyperparams}.

    \item \textbf{DL-HD + Covariates:} This is an extension of the above model where we use past $d$ days of precipitation data ($3 \leq d \leq 20$) from IMD, and past 3 days of reanalysis data. As stated earlier, the additional covariates include wind speed, temperature, soil moisture, cloud cover, vorticity, humidity, and divergence. The choice of these variables is justified later in Section \ref{sec:data}\ref{subsec:feature_selection}. The number of past days for reanalysis data were fixed to 3 as we did not observe any significant reduction in errors when incorporating information beyond 3 days. The input dimensions are now structured as $n \times d \times v$, where $v$ is the number of covariates. The output is again $n$-dimensional, forecasting the rainfall value at each grid. 
    
    \item \textbf{NWP:} HRES-NWP daily forecasts are available at $0.25^\circ$ resolution for lead times of 1 and 3 days. These are compared directly in case of IMD data at $0.25^\circ$, and compared after linear interpolation in the case of IMD data at $1^\circ$. NCEP-NWP forecasts are available at $1^\circ$ resolution on nearly all alternate days within the test period, and we make comparisons on the days when these forecasts are available. For each grid in the IMD dataset, we identify the best match within the re-aligned set and its four adjacent grids based on the criterion of the lowest forecast error. The forecast error is computed for each candidate grid, and the one with the minimum error is taken as the best matching grid. The forecast associated with this identified grid is considered the NCEP-NWP forecast for that specific location.

    \item \textbf{NWP+:} We combine the NWP forecasts at the target grid and the 4 neighbouring grids using a deep neural network, which is trained to minimize the error between the forecast and IMD ground truth for the particular grid. The resulting forecast is called NWP+ prediction. 
    The model here is trained from 2011 to 2020 and test forecasts are generated for 2021 - 2023. While using 4 surrounding grids improves the forecasts somewhat, we did not see further improvement with a higher number of grids.

    \item \textbf{Ensemble:} We combine the DL-HD + Covariates forecasts, and the HRES and NCEP forecasts of the 5 grids, to generate an ensemble forecast for each grid. This is done using a deep neural network. The models here are trained from 2011-2020, and forecasts are generated for 2021-2023.

\end{enumerate}
We also use the following simpler baselines to benchmark our models against:
\begin{enumerate}
    \item \textbf{Persistence:} This is a naive forecast which estimates the rainfall on day $d+1$ and the average of rainfall in days $d+1, d+2$ and $d+3$ as the observed rainfall on day $d$ for each grid. This is reported for the period 2012-2023.

    \item \textbf{Climatological mean:} This baseline estimates rainfall by computing the mean rainfall for the same calendar day across all previous years (1901--2011). It captures long-term seasonal trends and is used to benchmark model performance against historical averages.

\item \textbf{Rolling mean (20 Days):} This baseline forecasts rainfall as the average rainfall over the preceding 20 days for each grid point. It captures recent trends and smooths short-term variability but does not incorporate spatial or seasonal context.

\item \textbf{AR(1) - Temporal linear model:} This autoregressive baseline uses only the previous day's rainfall value at each grid point to forecast the next day's rainfall. A simple linear regression model is fit using this single temporal lag as input.

\item \textbf{AR(5) + spatial grids - spatiotemporal linear model:} This extends the autoregressive approach by incorporating the past 5 days of rainfall data for each grid point along with its four immediate neighboring grid points (up, down, left, right). A linear regression model is trained using this spatiotemporal input.

\end{enumerate}

\subsection{Loss function}

\begin{figure}[htbp]
  \centering
  \begin{subfigure}[b]{0.45\textwidth}
    \centering
    \includegraphics[width=\textwidth]{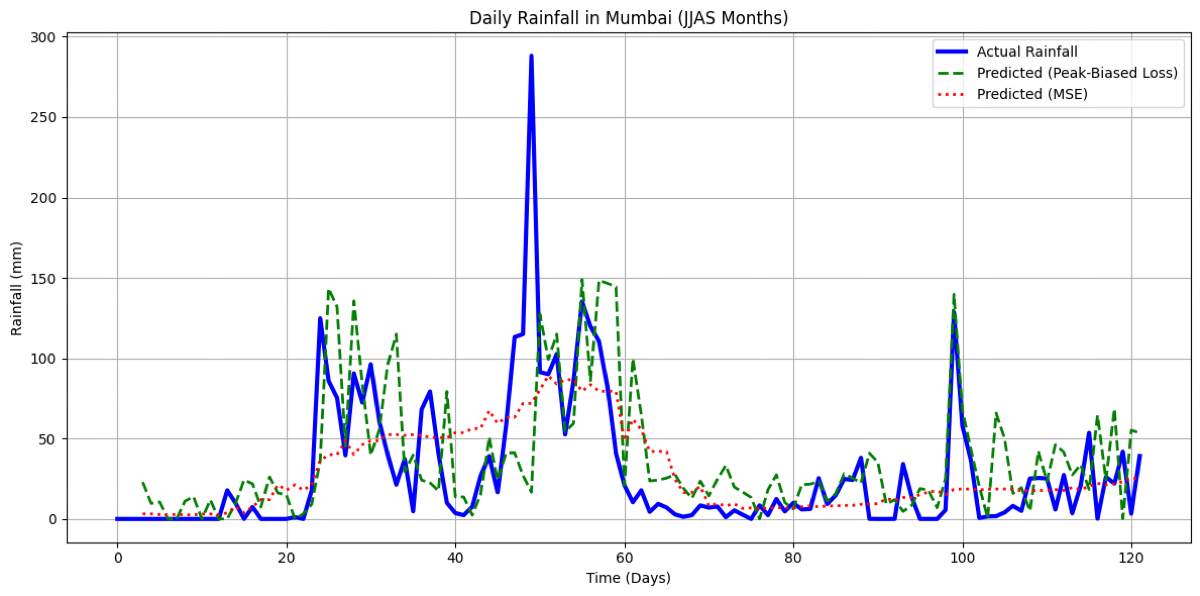}
    \caption{}
    \label{fig:mumbai_mse}
  \end{subfigure}
  \hfill
  \begin{subfigure}[b]{0.45\textwidth}
    \centering
    \includegraphics[width=\textwidth]{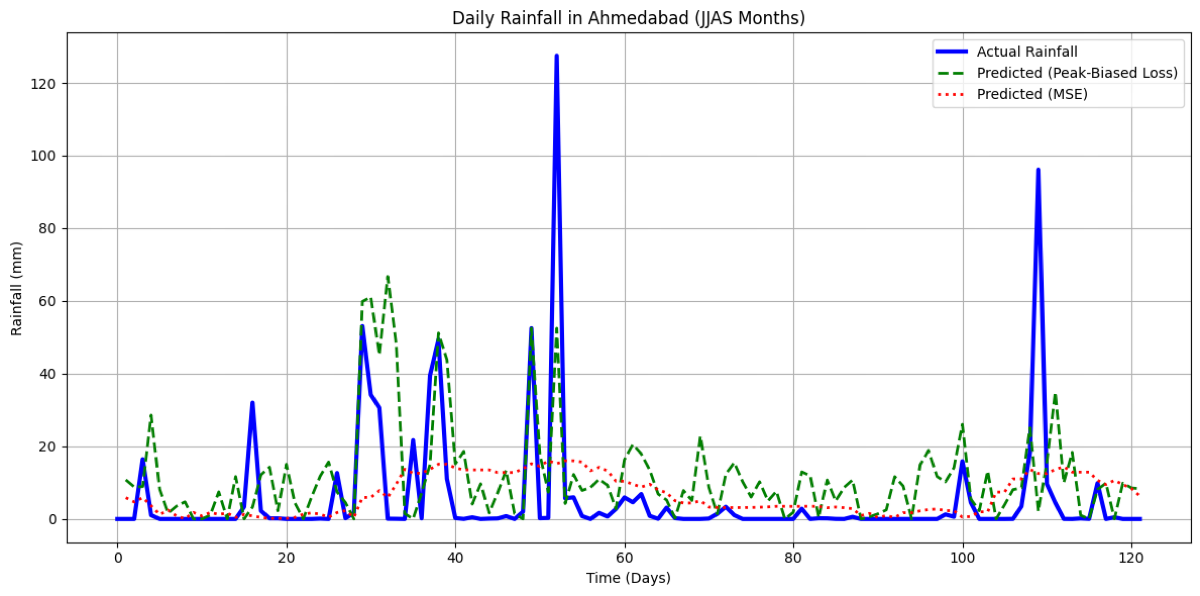}
    \caption{}
    \label{fig:ahm_mse}
  \end{subfigure}
  \caption{Plots comparing the predictions generated by the same Autoformer model under MSE and the proposed \emph{peak-biased loss} in (a) Mumbai and (b) Ahmedabad respectively.}
  \label{fig:fig10_comparison}
\end{figure}

Since rainfall forecasting is a regression task, the mean squared error (MSE) is the conventional choice of loss function. However, during training, we found that models optimized with MSE tend to produce overly smooth predictions that fail to capture extreme rainfall events. This behavior is particularly evident in Figures~\ref{fig:mumbai_mse} and~\ref{fig:ahm_mse}, where MSE-trained models (red dotted line) track the long-term average but significantly underestimate peaks for Mumbai and Ahmedabad, respectively. This limitation stems from MSE’s symmetric treatment of over- and underestimation errors, which biases the model toward minimizing large deviations without prioritizing rare but high-impact events.\\
To address this, we propose a \emph{peak-biased loss function} that places greater emphasis on underestimation. This reflects the practical importance of accurately forecasting extreme rainfall, where missing a peak can have far more serious consequences than a false alarm. The loss function is defined as:

\begin{equation}
\label{eq:pbl}
L = \frac{1}{N} \sum_{t=1}^{N} \left[ \mathbb{I}(\hat{r}_t < r_t) \cdot |r_t - \hat{r}_t|^{\alpha} + \mathbb{I}(\hat{r}_t > r_t) \cdot |\hat{r}_t - r_t|^{\beta} \right],
\end{equation}

\noindent where $r_t$ and $\hat{r}_t$ denote the observed and predicted rainfall at time $t$, respectively, and $\alpha > \beta$ ensures that underestimation is penalized more heavily than overestimation. We use $\alpha = 1.5$, $\beta = 1.0$, selected empirically to optimize forecast quality.\\
To validate this choice, we conduct a sensitivity analysis across various values of $\alpha$ and $\beta$. Results across all percentiles (Tables~\ref{tab:confmat_0mm}--\ref{tab:confmat_26mm}) consistently show $\alpha = 1.5$, $\beta = 1.0$ achieving optimal balance between detection accuracy and false alarm rates.\\
This design is also supported by prior work addressing rare event prediction. For instance, \cite{shi2017deep} and \cite{xu2024extremecast} use such custom weighted loss functions to improve performance on infrequent but critical events.
\noindent
\\The advantage of our loss function is also visually evident. In the case of Mumbai (Figure~\ref{fig:mumbai_mse}), the model trained with the peak-biased loss (green dashed line) captures multiple significant rainfall events missed by the MSE-trained model, particularly between days 20-40 and near day 50. In Ahmedabad (Figure~\ref{fig:ahm_mse}), the peak-biased model estimates the $\sim$130mm rainfall at day 50 more accurately, while the MSE model significantly underpredicts it.\\
For completeness, we report both the proposed peak-biased loss $L$ and MSE on the test set (Tables~\ref{tab:loss_india_1day_0.25deg} and~\ref{tab:loss_india_3day_0.25}). The trends are consistent across both metrics: lower MSE corresponds to lower $L$, with peak-biased models outperforming in both average error and peak detection.

\subsection{Feature selection}\label{subsec:feature_selection}
We adopt a sequential feature selection approach to identify input variables for forecasting rainfall. This greedy algorithm adds one variable at a time, selecting at each step the variable that yields the greatest reduction in forecasting error, as measured by the peak-biased loss function.\\
The process begins by evaluating each variable individually. The variable with the lowest standalone error is selected first. At each subsequent step, the model assesses the marginal improvement from adding each remaining variable to the current set and selects the one that most improves performance. This continues until no further variable leads to a significant reduction in error.
\\Table \ref{tab:feature_selection} outlines the progression of this selection. Past precipitation is chosen first, achieving the lowest initial error. It is followed by cloud cover, vorticity, humidity, soil moisture, wind, and finally temperature, each selected for their incremental contribution to minimizing the loss. This method prioritizes immediate gains in predictive accuracy rather than exploring all possible feature combinations.

\begin{table}[ht]
\caption{Peak-biased loss ($mm^{1.5}+ mm$) recorded for different combinations of input variables using a greedy approach}
    \label{tab:feature_selection}
    \centering
    \scriptsize
    \begin{tabular}{|l|c|c|c|c|c|c|c|}
        \hline
        \textbf{Variables} & \textbf{Cloud Cover} & \textbf{H Wind} & \textbf{V Wind} & \textbf{Temperature} & \textbf{Humidity} & \textbf{Soil Moisture} & \textbf{Vorticity} \\
        \textbf{} & \textbf{(CC)} & \textbf{(HW)} & \textbf{(VW)} & \textbf{(T)} & \textbf{(H)} & \textbf{(SM)} & \textbf{(Vo)} \\
        \hline
        Precipitation (Ppt) & \textbf{19.95} & 21.13 & 21.48 & 21.01 & 20.71 & 20.52 & 20.39 \\
        Ppt + CC  & - & 19.63 & 19.88 & 20.20 & 20.01 & \textbf{19.38} & 19.47 \\
        Ppt + CC + Vo  & - & 19.73 & 19.71 & 19.36 & \textbf{19.07} & 19.33 & - \\
        Ppt + CC + Vo + H & - & 18.91 & 18.96 & 18.99 & - & \textbf{18.56} & - \\
        Ppt + CC + Vo + H + SM & - & \textbf{18.42} & 18.59 & 18.53 & - & - & - \\
        \hline
    \end{tabular}
    
\end{table}
\subsection{Model Configuration and Training}
All experiments are conducted in Python, utilizing the TensorFlow and Keras libraries. DL-HD and DL-HD+Covariates forecasts are made using the Autoformer architecture. This is a  recurrent model and we use past $d$ days of $v$ variables as input, where $d$ ranges from 3 to 20. The number of parameters depend roughly linearly on $d$, with the number of parameters being approximately 200M for $d=12$. This architecture is adopted from \cite{wu2021autoformer}. Since there are fewer data points, the models for NWP+ and Ensemble are trained using smaller feed-forward neural networks, with 2 hidden layers. All models are trained using the Adam optimizer, optimizing the peak-biased loss specified in  \eqref{eq:pbl}. More details on the architecture of these models are given in the Appendix~\ref{app:nn_hyperparams}. To ensure robustness of our models, each experiment is conducted across 10 independent runs, employing randomly generated seeds to initialize neural network parameters differently. Performance metrics are reported on the average prediction obtained from these 10 runs.

\section{Results}
The predictions based on the models specified in Section~\ref{sec:data}\ref{section:models} are compared with the ground truth daily rainfall data from IMD. We outline results both in an all-India average sense and separately for a set of key cities to capture regional performance differences. Below we present results corresponding to IMD data at $0.25^\circ \times 0.25^\circ$ resolution. The results for $1^\circ \times 1^\circ$ IMD data are given in Appendix~\ref{app:1deg}.
\subsection{Comparison for entire India} 

\begin{figure}[htp]
    \centering
    
    \textbf{Spatial distribution of 1-day forecasts for 15 July 2022}
    \vspace{0.3em}

    \begin{subfigure}{0.3\textwidth}
        \includegraphics[width=\linewidth]{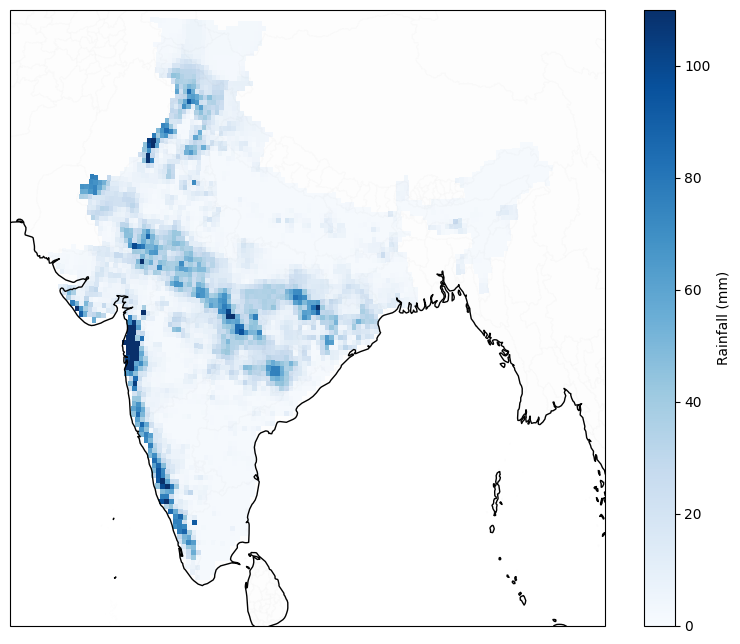}
        \caption{IMD Ground Truth}
        \label{fig:1day_subfig1}
    \end{subfigure}
    \hfill
    \begin{subfigure}{0.3\textwidth}
        \includegraphics[width=\linewidth]{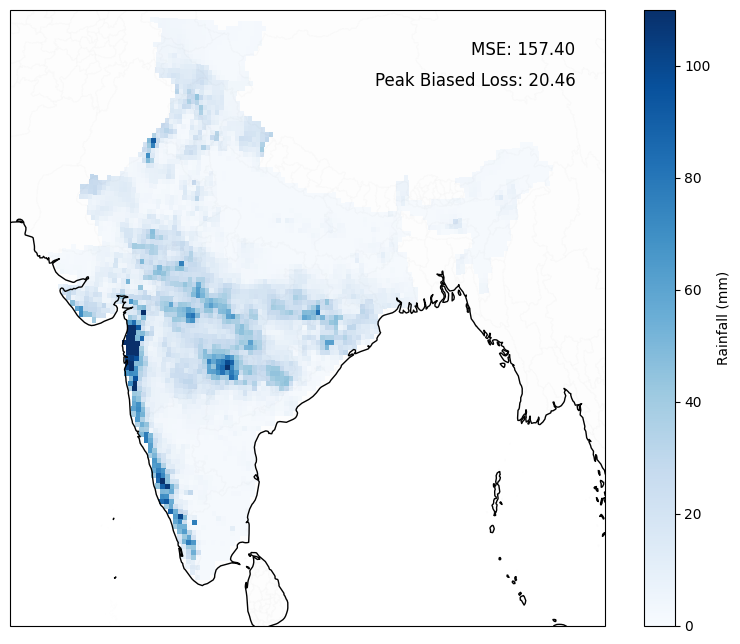}
        \caption{DL-HD+Covariates}
        \label{fig:1day_subfig2}
    \end{subfigure}
    \hfill
    \begin{subfigure}{0.3\textwidth}
        \includegraphics[width=\linewidth]{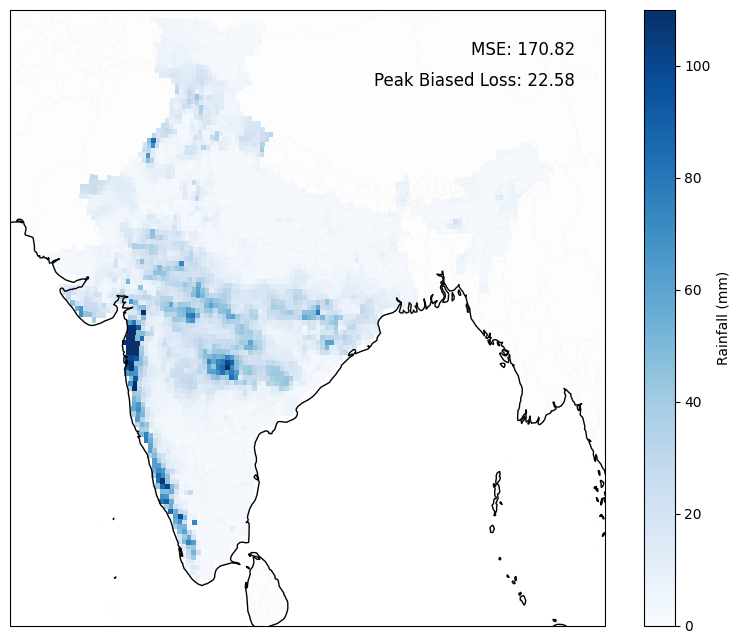}
        \caption{DL-HD}
        \label{fig:1day_subfig3}
    \end{subfigure}

    \vspace{0.5em}
    
    \begin{subfigure}{0.3\textwidth}
        \includegraphics[width=\linewidth]{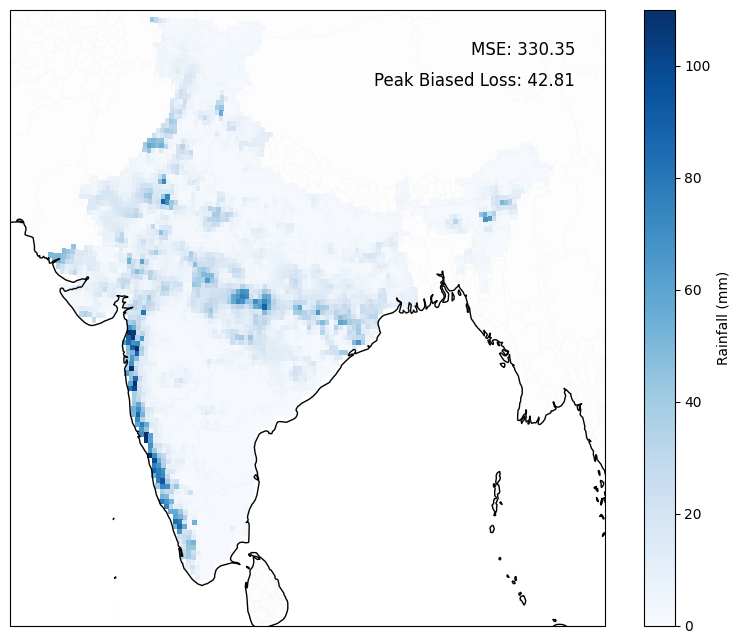}
        \caption{HRES}
        \label{fig:1day_subfig4}
    \end{subfigure}
    \hfill
    \begin{subfigure}{0.3\textwidth}
        \includegraphics[width=\linewidth]{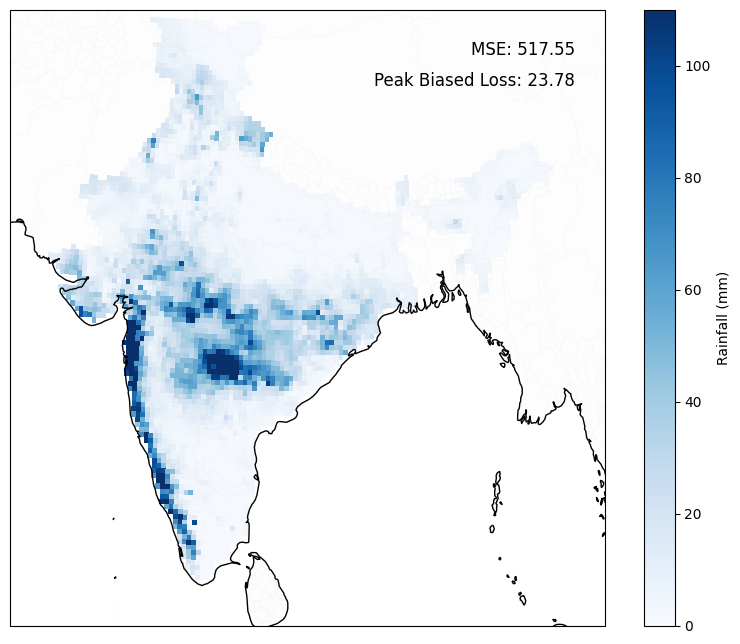}
        \caption{HRES-NWP+}
        \label{fig:1day_subfig5}
    \end{subfigure}
    \hfill
    \begin{subfigure}{0.3\textwidth}
        \includegraphics[width=\linewidth]{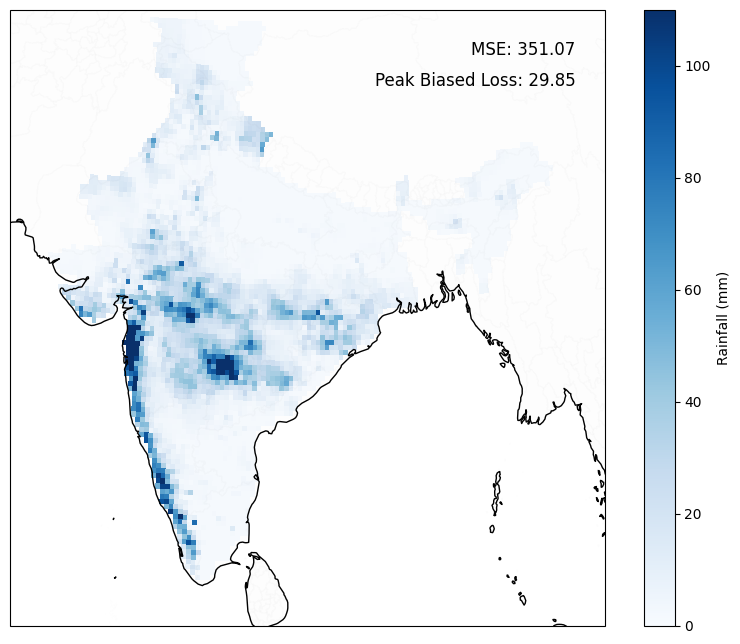}
        \caption{Ensemble}
        \label{fig:1day_subfig6}
    \end{subfigure}

    \vspace{1em}
    
    \textbf{Spatial distribution of 3-day forecasts for 15 July 2022}
    \vspace{0.3em}

    \begin{subfigure}{0.3\textwidth}
        \includegraphics[width=\linewidth]{gt_15jul22.png}
        \caption{IMD Ground Truth}
        \label{fig:3day_subfig1}
    \end{subfigure}
    \hfill
    \begin{subfigure}{0.3\textwidth}
        \includegraphics[width=\linewidth]{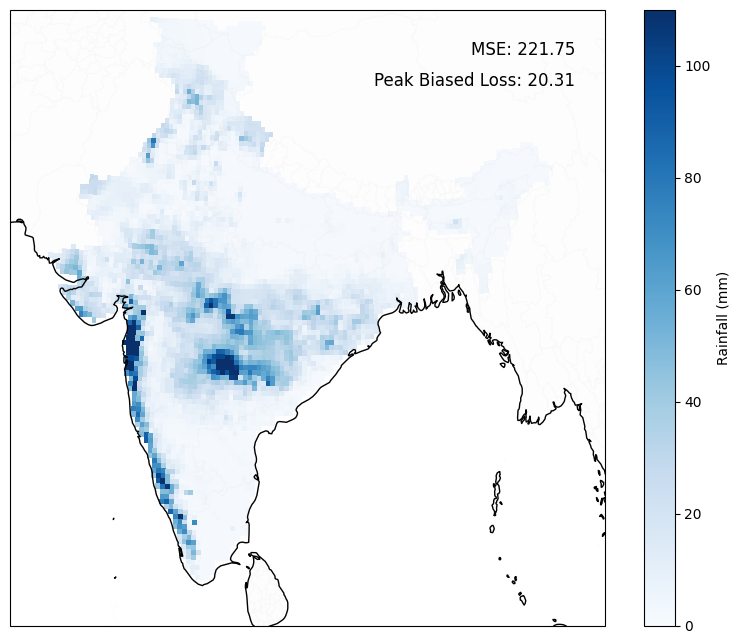}
        \caption{DL-HD+Covariates}
        \label{fig:3day_subfig2}
    \end{subfigure}
    \hfill
    \begin{subfigure}{0.3\textwidth}
        \includegraphics[width=\linewidth]{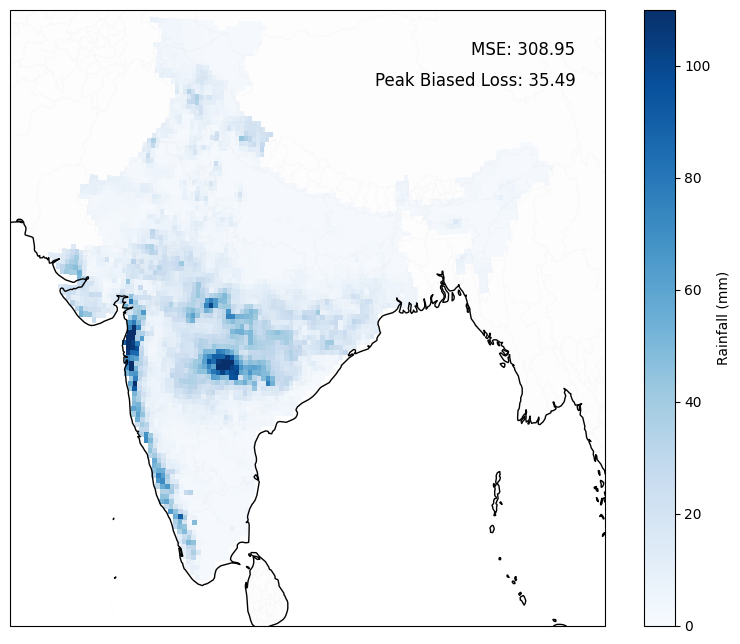}
        \caption{DL-HD}
        \label{fig:3day_subfig3}
    \end{subfigure}

    \vspace{0.5em}
    
    \begin{subfigure}{0.3\textwidth}
        \includegraphics[width=\linewidth]{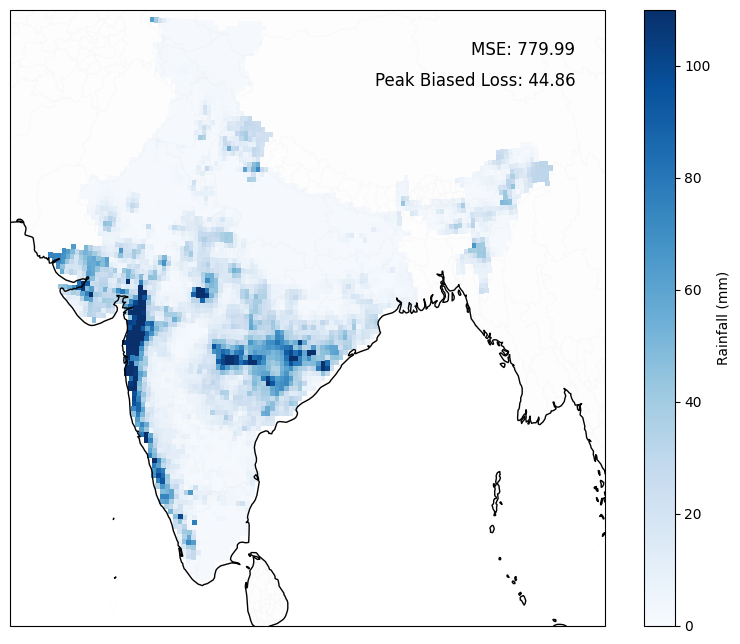}
        \caption{HRES}
        \label{fig:3day_subfig4}
    \end{subfigure}
    \hfill
    \begin{subfigure}{0.3\textwidth}
        \includegraphics[width=\linewidth]{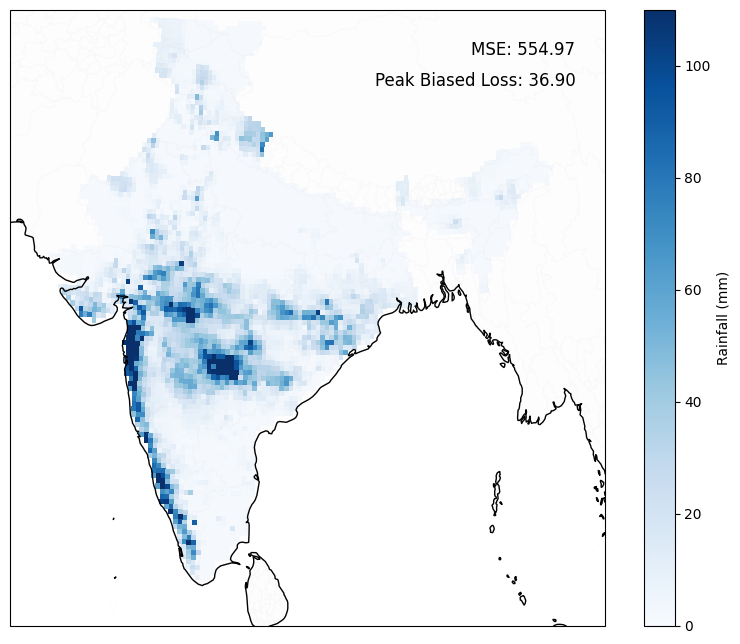}
        \caption{HRES-NWP+}
        \label{fig:3day_subfig5}
    \end{subfigure}
    \hfill
    \begin{subfigure}{0.3\textwidth}
        \includegraphics[width=\linewidth]{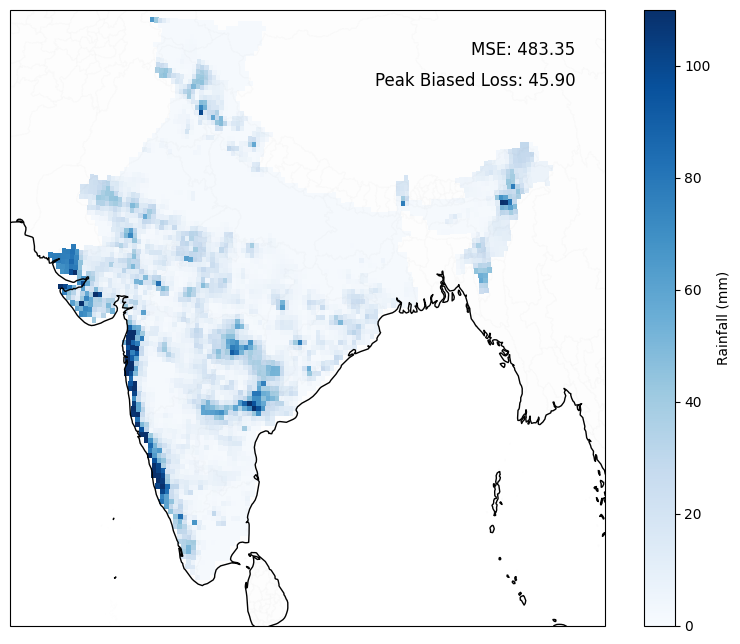}
        \caption{Ensemble}
        \label{fig:3day_subfig6}
    \end{subfigure}

    \caption{Spatial distribution of forecasts for 15 July 2022. Top: 1-day forecasts. Bottom: 3-day forecasts.}
    \label{fig:combined_forecasts_2022_07_15}
\end{figure}

The average peak-biased loss over India for 1-day forecasts is presented in Table~\ref{tab:loss_india_1day_0.25deg}. We also compare the spatial distribution of prediction skill for 1 and 3
day prediction on July 15, 2022 in Figure \ref{fig:combined_forecasts_2022_07_15}. From the analysis, we make the following observations:

\begin{enumerate}
    \item \textbf{DL-HD + Covariates} achieves the lowest error and outperforms all other models.
    
    \item Among the climatological baselines, the \textbf{Rolling Mean (20 Days)} performs worst, exhibiting 81.68\% higher error than DL-HD + Covariates, followed by the \textbf{Climatological Mean} with 59.02\% higher error.
    
    \item \textbf{Persistence} has relatively high error but performs better than the climatological baselines, with 39.36\% higher MSE.
    
    \item \textbf{HRES-NWP} performs better than persistence but still has 32.90\% higher error relative to DL-HD + Covariates.
    
    \item Pooling neighboring grids in \textbf{HRES-NWP+} slightly improves over HRES, yet remains 28.21\% worse than DL-HD + Covariates.
    
    \item The \textbf{Ensemble (NWP + DL-HD + Covariates)} demonstrates improvement over NWP alone, with an error 9.56\% higher than DL-HD + Covariates.
    
    \item The regression-based models, \textbf{AR(1)} and \textbf{AR(5) + Spatial Grids}, perform worse than DL-HD + Covariates, with MSEs 32.18\% and 30.98\% higher respectively, indicating the DL model's advantage in capturing non-linear patterns beyond lag-based predictors.
\end{enumerate}

\subsection{Comparison for key cities} 
\begin{figure}[h]
    \centering
    \includegraphics[width=0.75\textwidth]{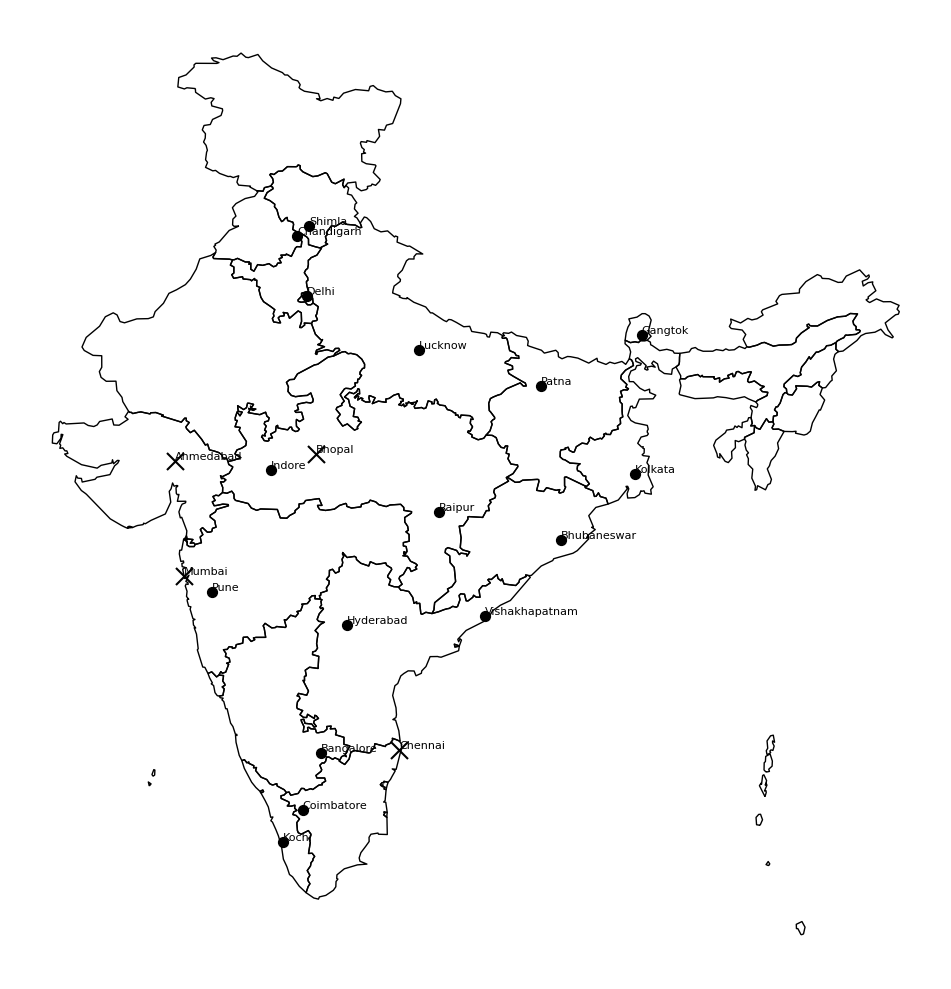}
    \caption{Study area map showing the selected 20 cities across India, including both coastal and landlocked locations. The cities marked with a cross are chosen for detailed analysis based on the rainfall during JJAS and the population in these cities.}
    \label{fig:study_area_map}
\end{figure}

We also analyzed the model performance separately for 20 of the most populated cities spread across India. These cities were selected to provide a representative distribution across coastal and landlocked regions, and to ensure significant rainfall during the monsoon months (JJAS), which is important for evaluating rainfall forecasting accuracy. A map showing the geographical distribution of these selected cities is included in Figure~\ref{fig:study_area_map}.

The average peak-biased loss for these cities is shown in Table~\ref{tab:loss_1day_cities_0.25deg} for 1-day forecasts, and in Table~\ref{tab:loss_3day_cities_0.25deg} for 3-day forecasts. The last row in both tables shows the excess error percentage of the different forecasts compared to the DL-HD+Covariates forecast. We make similar observations here as for the whole of India.

\begin{figure}
  \caption{1-day forecasts for Mumbai in July and August 2022. DL-HD+Covariates predictions closely track the ground truth, while HRES predictions tend to over estimate the rainfall. The ensemble is a significant improvement over NWP alone, and can be seen to capture most of the high rainfall events during this period.}
  \label{fig:Mumbai_1day_cities_loss}
  \centering
  \begin{subfigure}[b]{0.4\textwidth}
    \includegraphics[width=\textwidth]{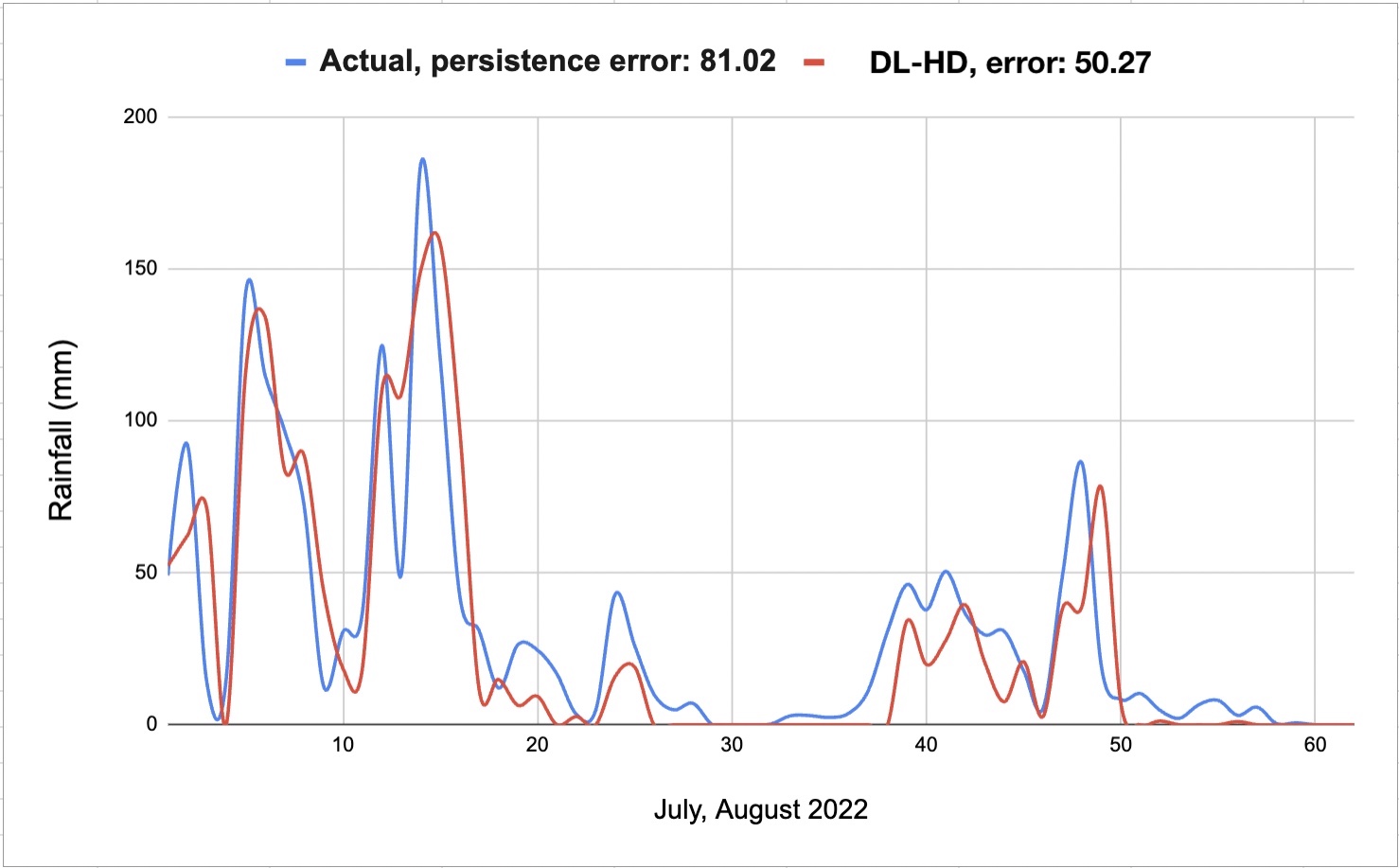}
    \caption{DL-HD + Covariates vs IMD}
  \end{subfigure}
  \begin{subfigure}[b]{0.4\textwidth}
    \includegraphics[width=\textwidth]{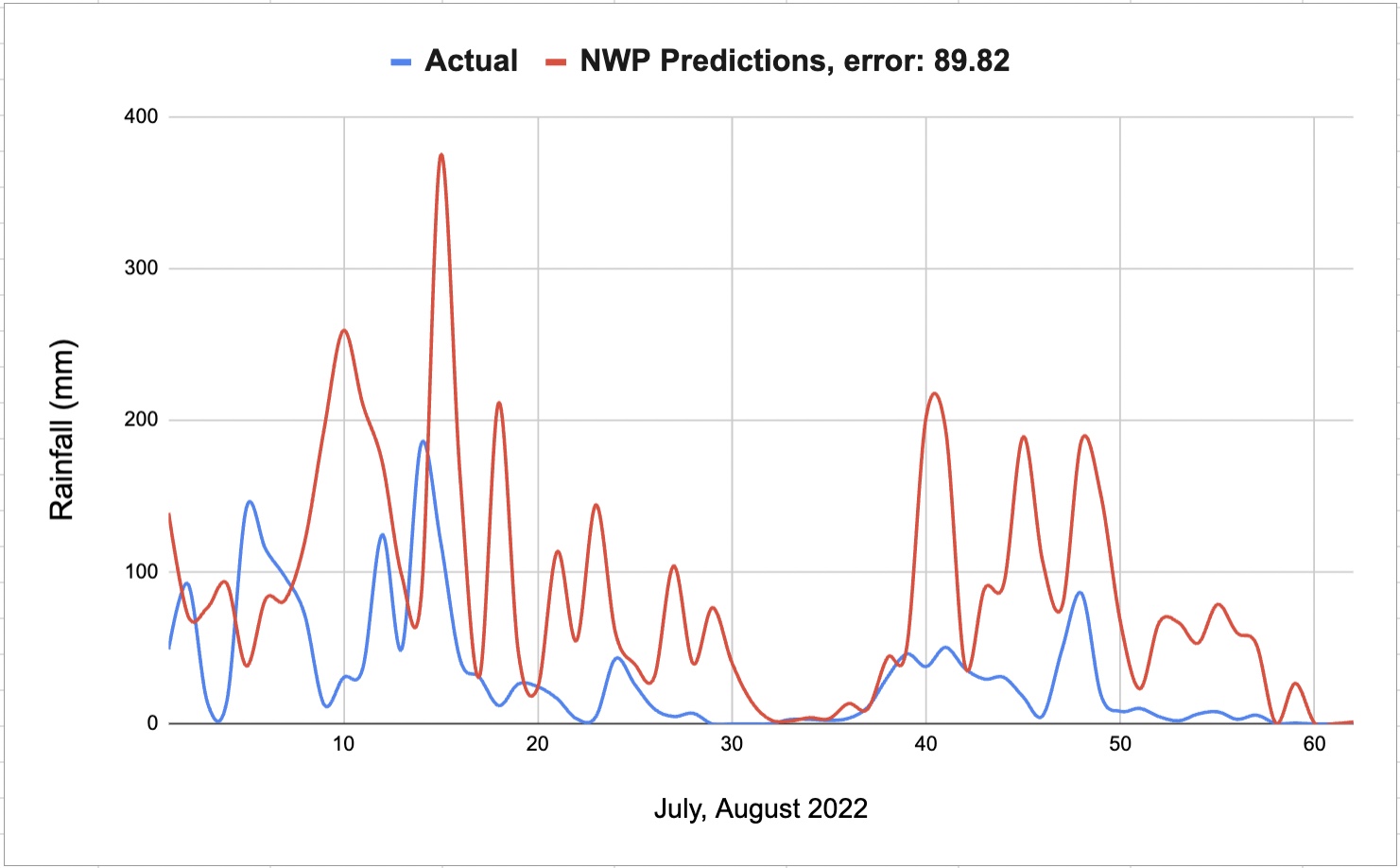}
    \caption{HRES vs IMD}
  \end{subfigure}
  
  \medskip
  
  \begin{subfigure}[b]{0.4\textwidth}
    \includegraphics[width=\textwidth]{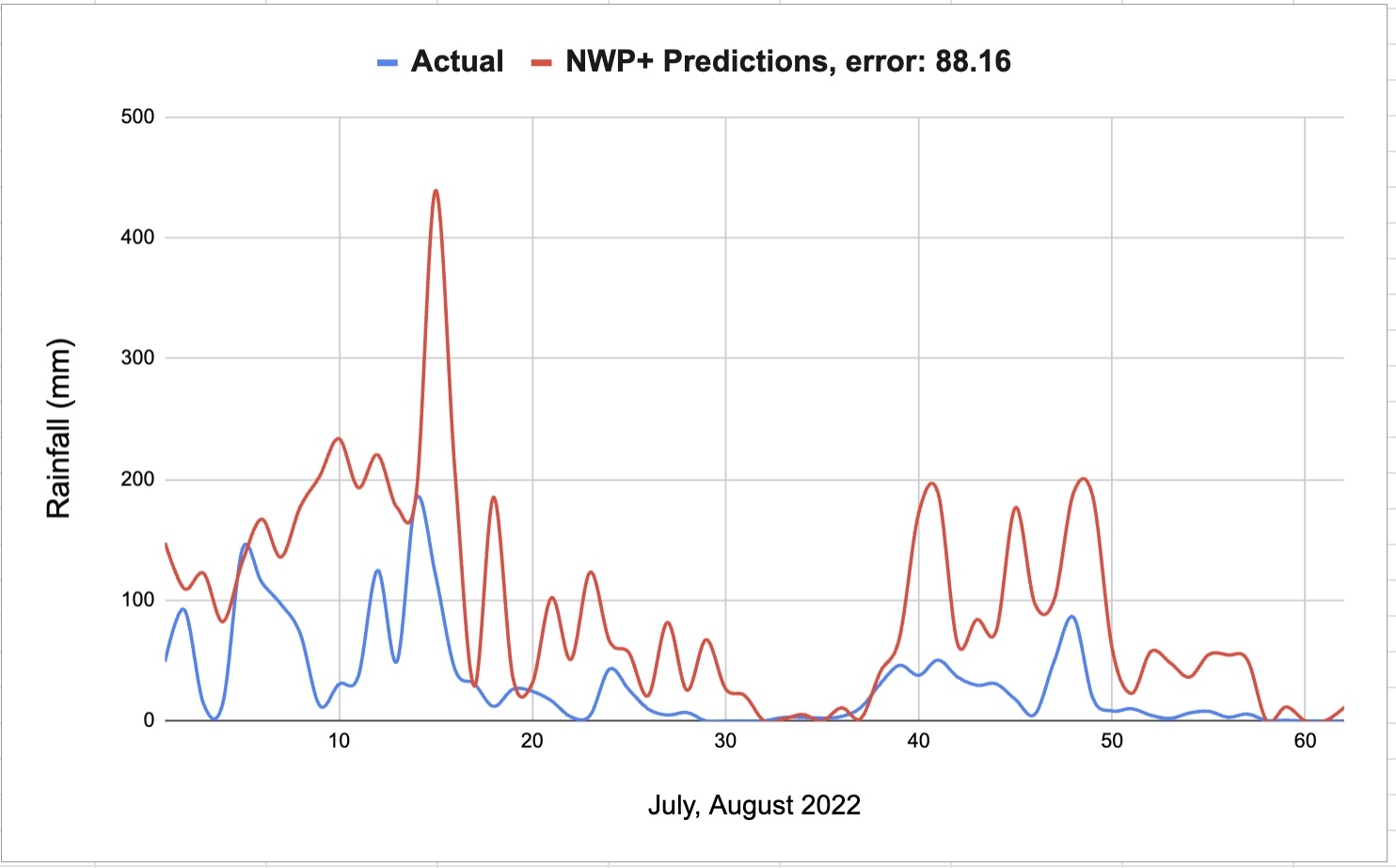}
    \caption{HRES-NWP+ vs IMD}
  \end{subfigure}
  \begin{subfigure}[b]{0.4\textwidth}
    \includegraphics[width=\textwidth]{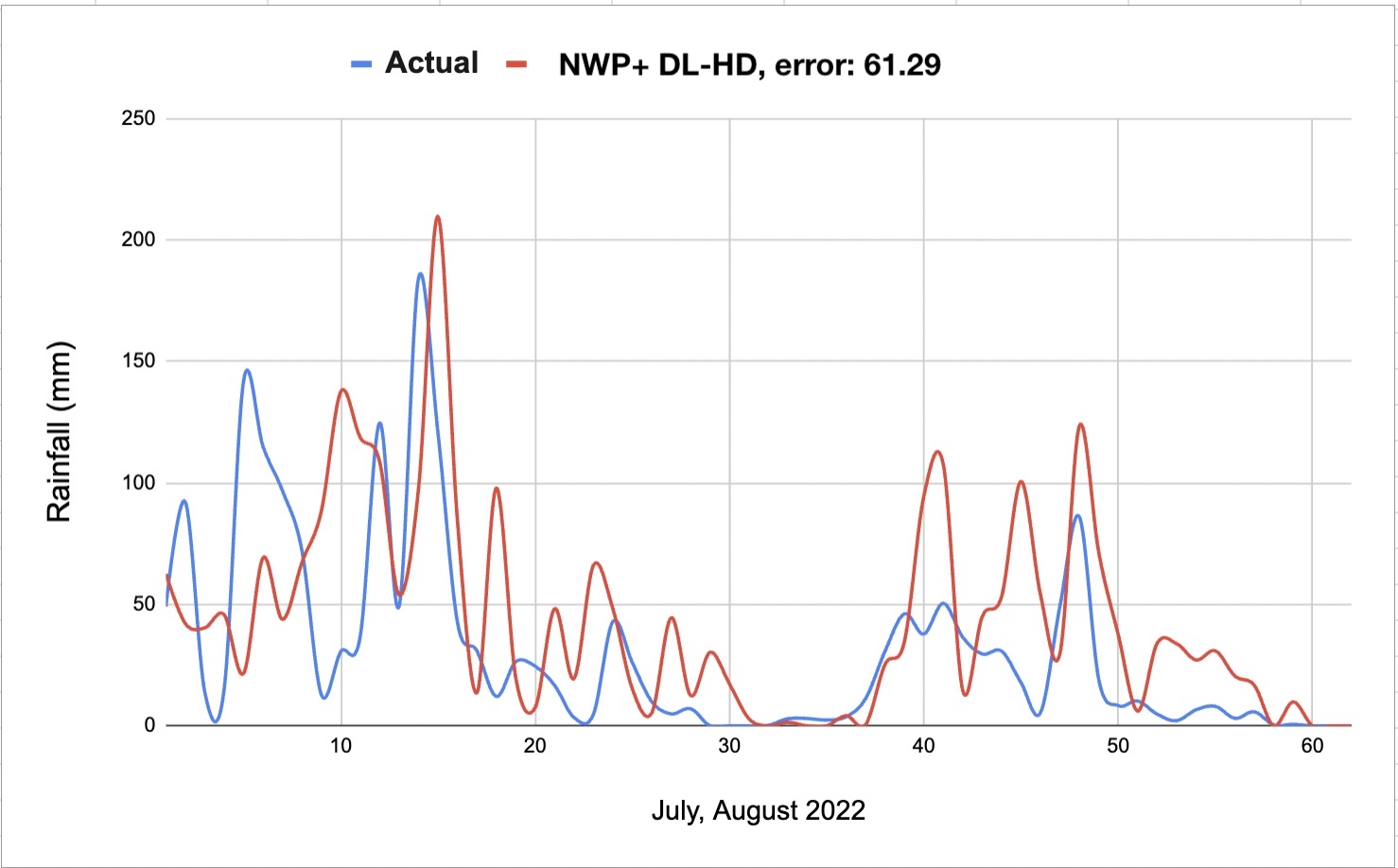}
    \caption{Ensemble vs IMD}
  \end{subfigure}

\end{figure}

\begin{figure}
  \caption{1-day forecasts for Ahmedabad in July and August 2022. DL-HD+Covariates closely track the ground truth, capturing most high rainfall events during this period. HRES predictions consistently overestimate the rainfall}
  \label{fig:Ahmedabad_1day_cities_loss}
  \centering
  \begin{subfigure}[b]{0.4\textwidth}
    \includegraphics[width=\textwidth]{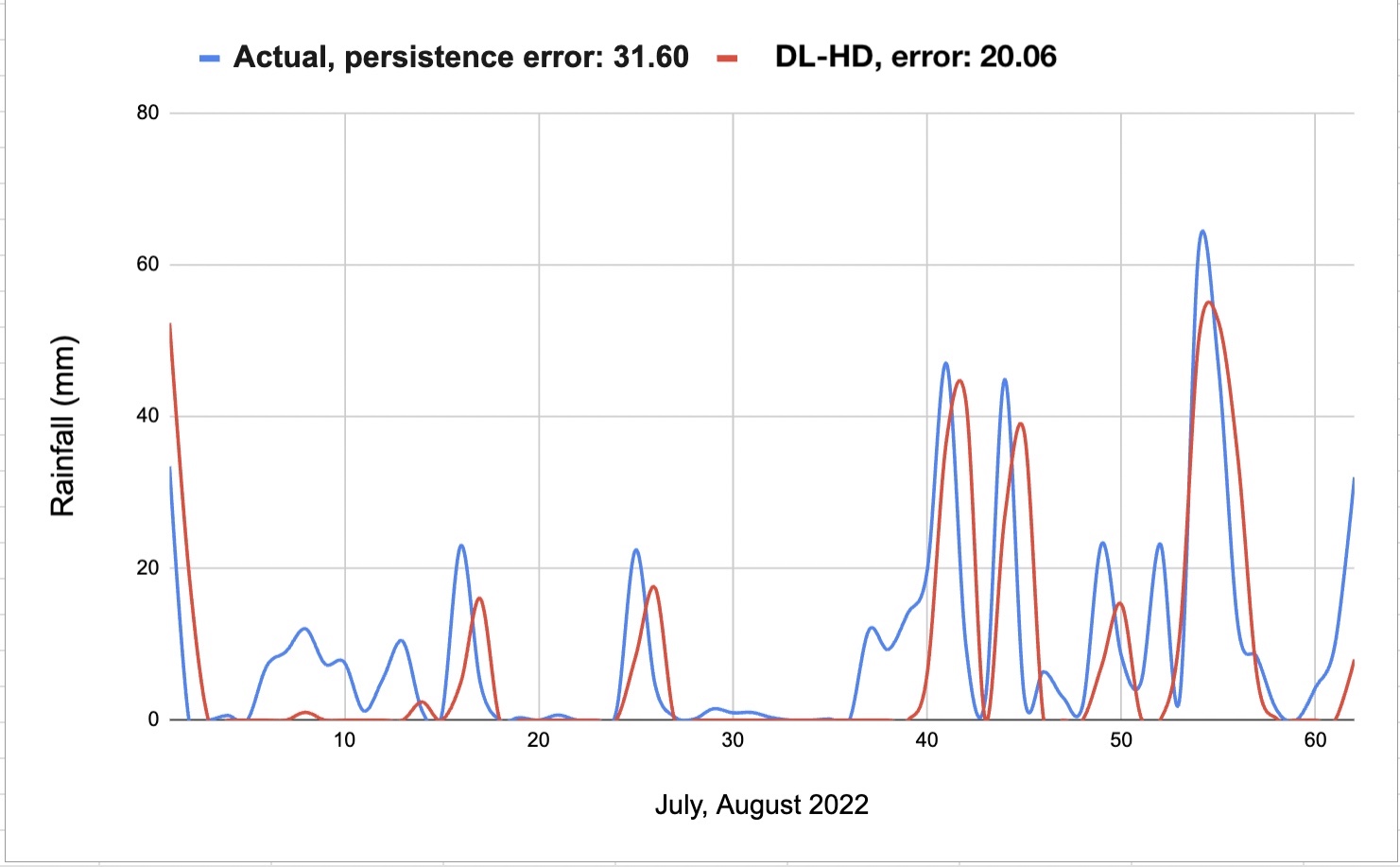}
    \caption{DL-HD + Covariates vs IMD}
  \end{subfigure}
  \begin{subfigure}[b]{0.4\textwidth}
    \includegraphics[width=\textwidth]{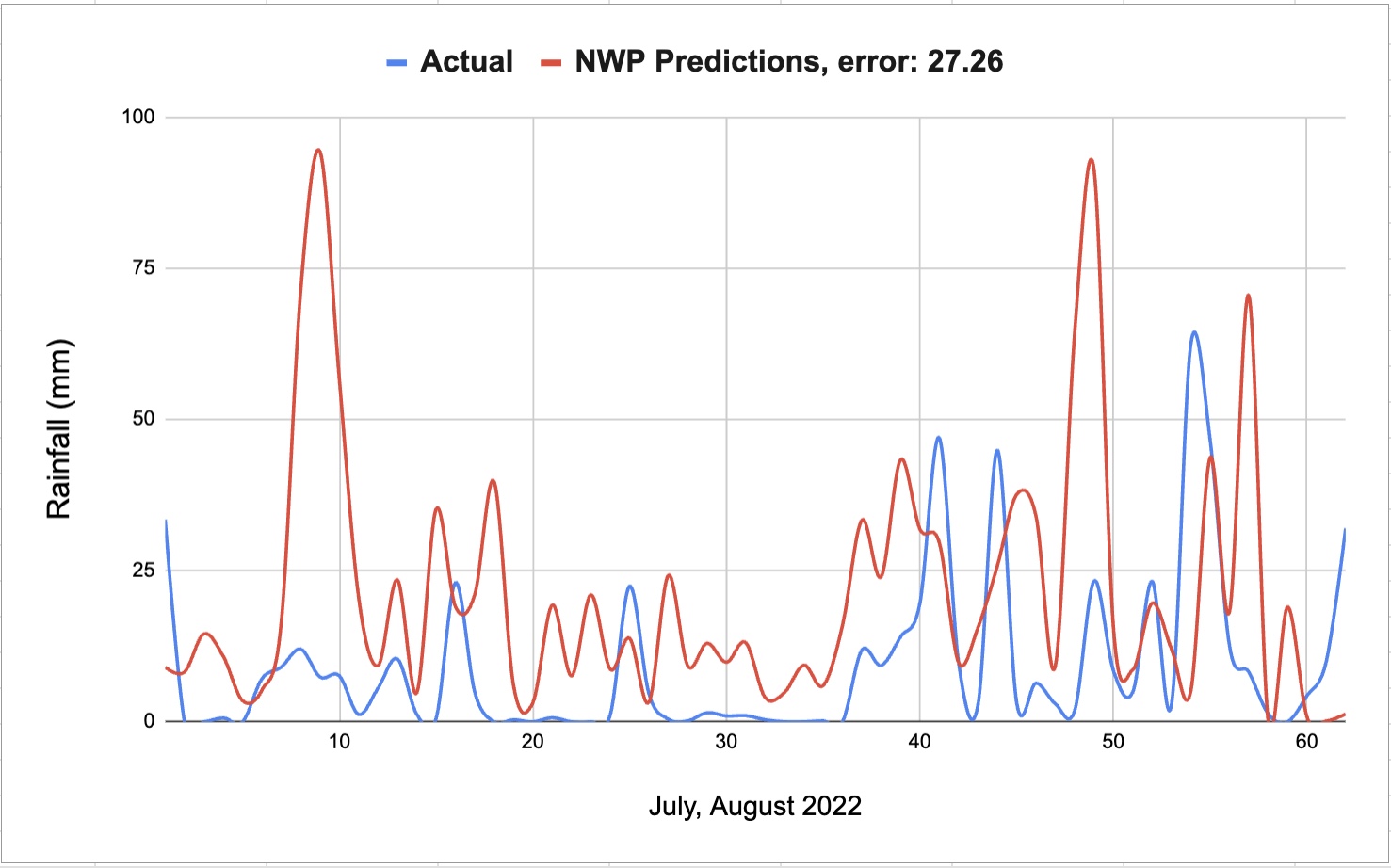}
    \caption{HRES vs IMD}
  \end{subfigure}
  
  \medskip
  
  \begin{subfigure}[b]{0.4\textwidth}
    \includegraphics[width=\textwidth]{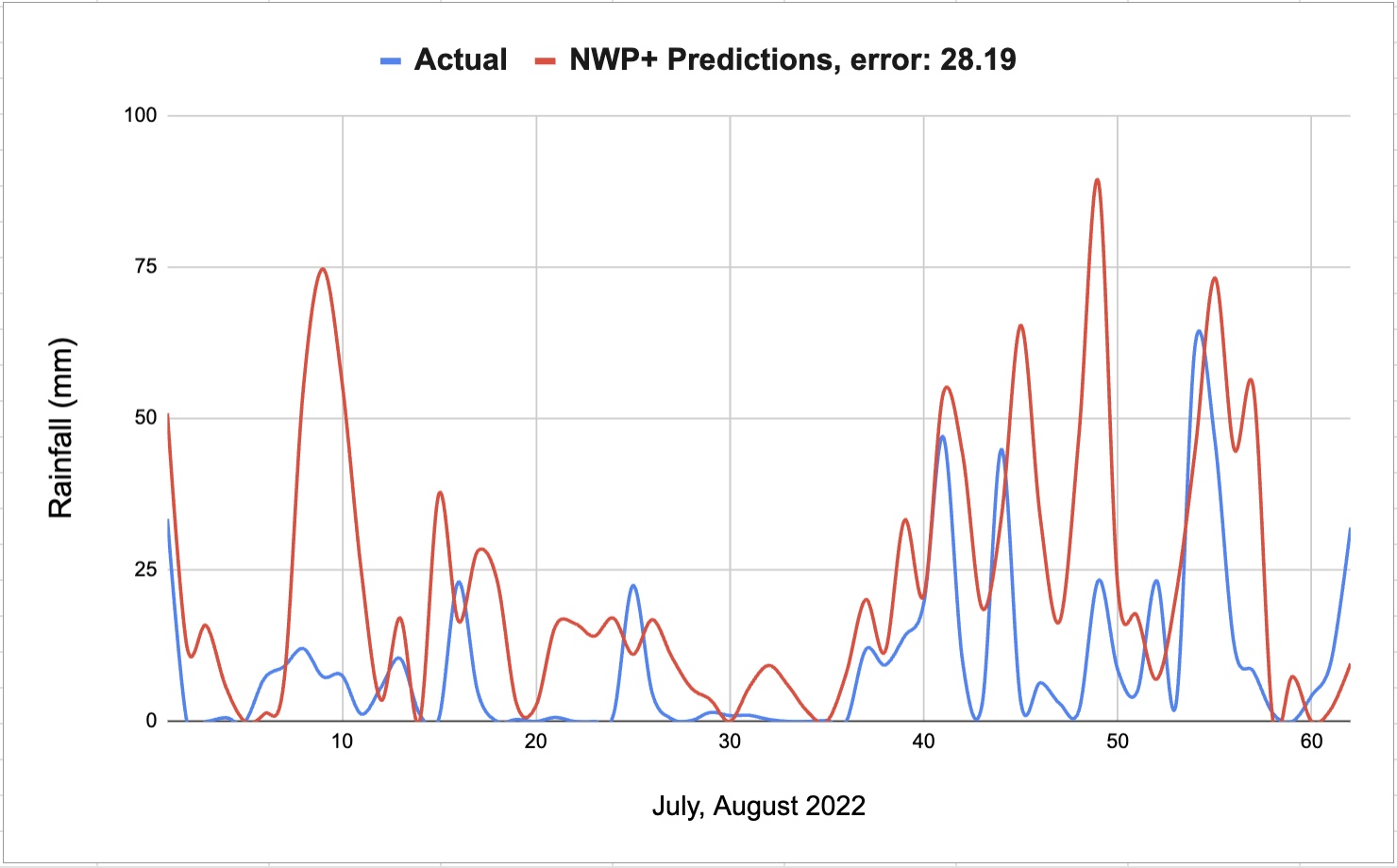}
    \caption{HRES-NWP+ vs IMD}
  \end{subfigure}
  \begin{subfigure}[b]{0.4\textwidth}
    \includegraphics[width=\textwidth]{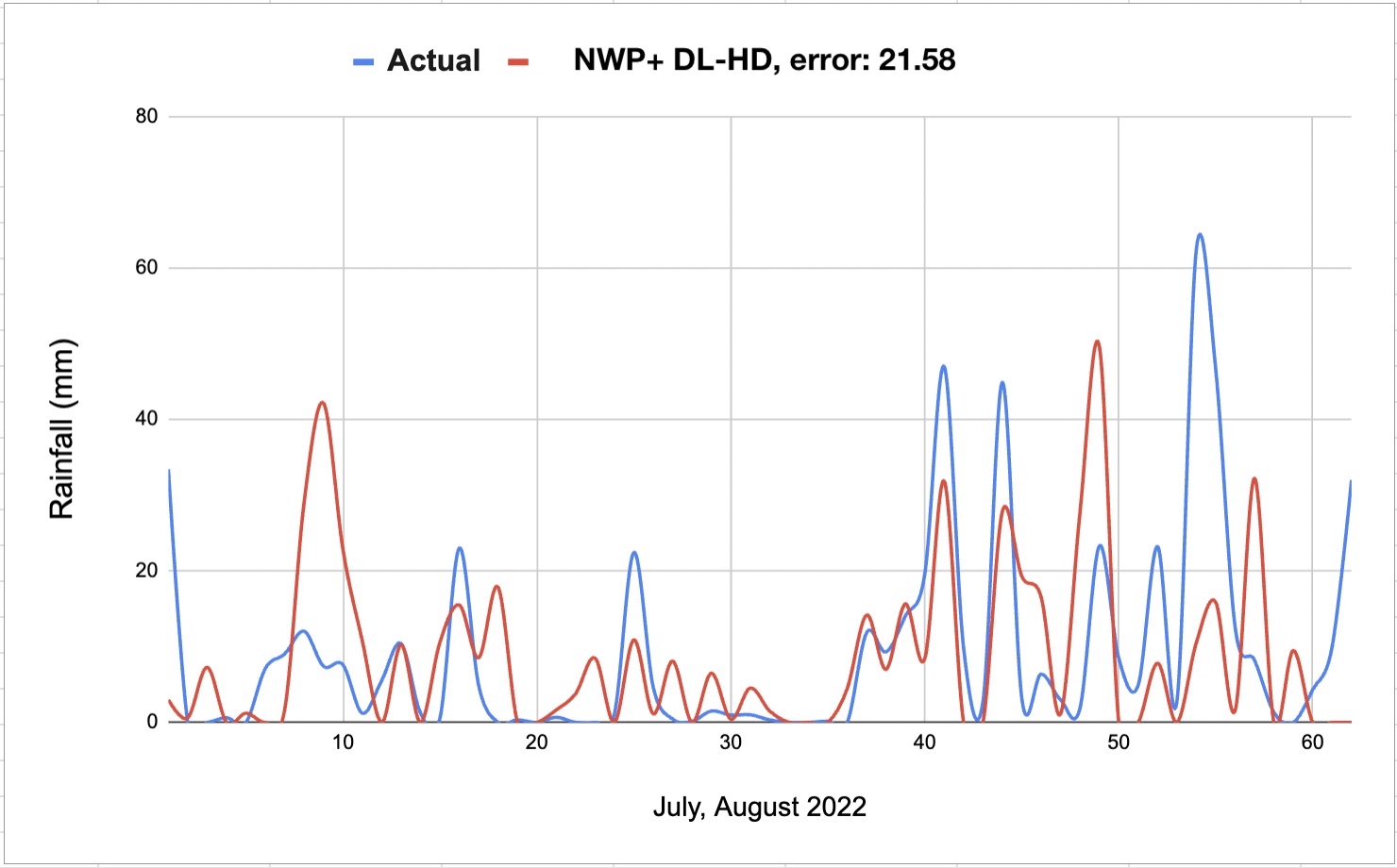}
    \caption{Ensemble vs IMD}
    
  \end{subfigure}

\end{figure}

\begin{figure}
  
  \caption{1-day forecasts for Chennai in July and August 2022. DL-HD+Covariates closely track the IMD ground truth. HRES predictions consistently overestimate the rainfall}
  \label{fig:Chennai_1day_cities_loss}
  \centering
  \begin{subfigure}[b]{0.4\textwidth}
    \includegraphics[width=\textwidth]{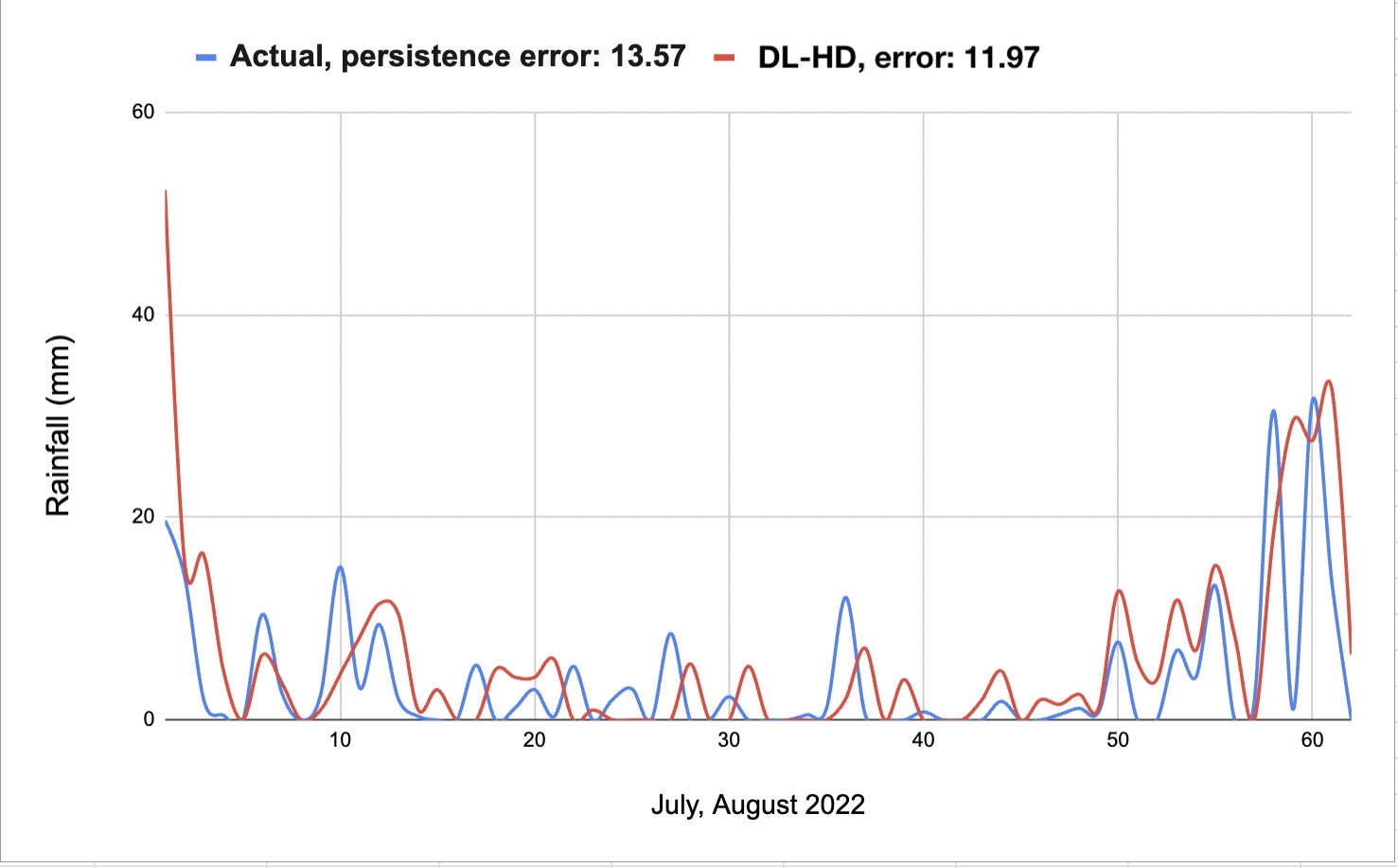}
    \caption{DL-HD + Covariates vs IMD}

  \end{subfigure}
  \begin{subfigure}[b]{0.4\textwidth}
    \includegraphics[width=\textwidth]{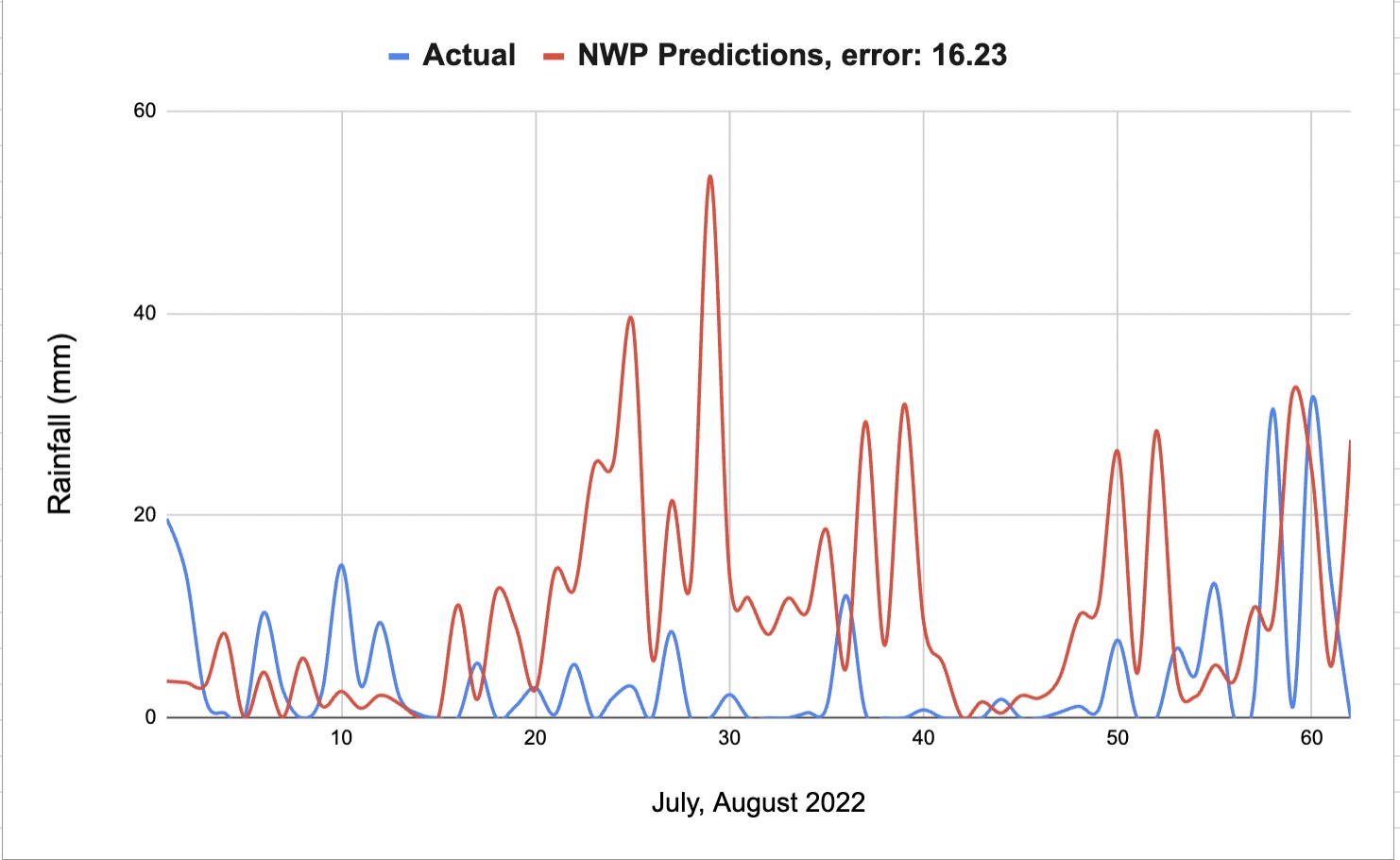}
    \caption{HRES vs IMD}

  \end{subfigure}
  
  \medskip
  
  \begin{subfigure}[b]{0.4\textwidth}
    \includegraphics[width=\textwidth]{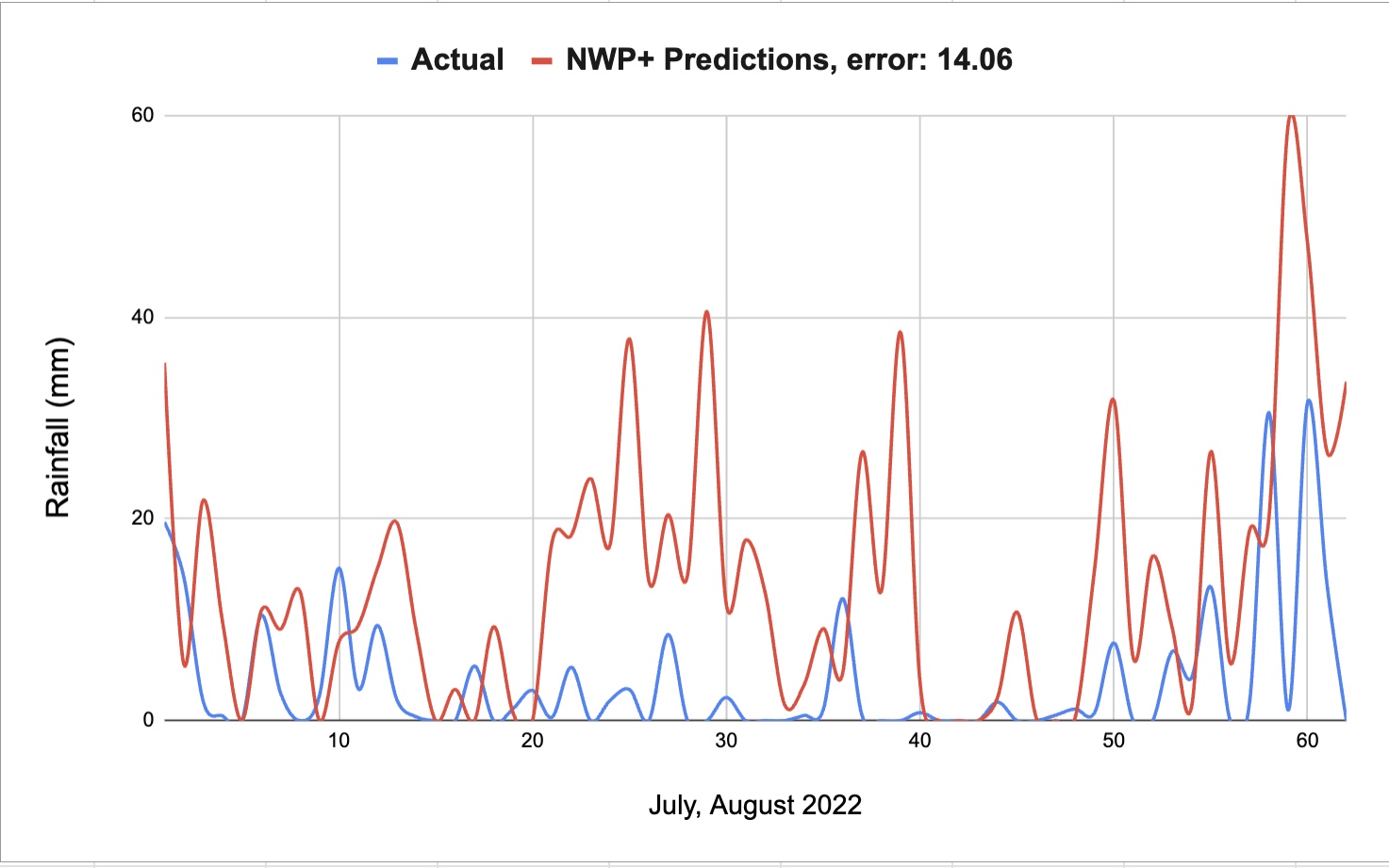}
    \caption{HRES-NWP+ vs IMD}

  \end{subfigure}
  \begin{subfigure}[b]{0.4\textwidth}
    \includegraphics[width=\textwidth]{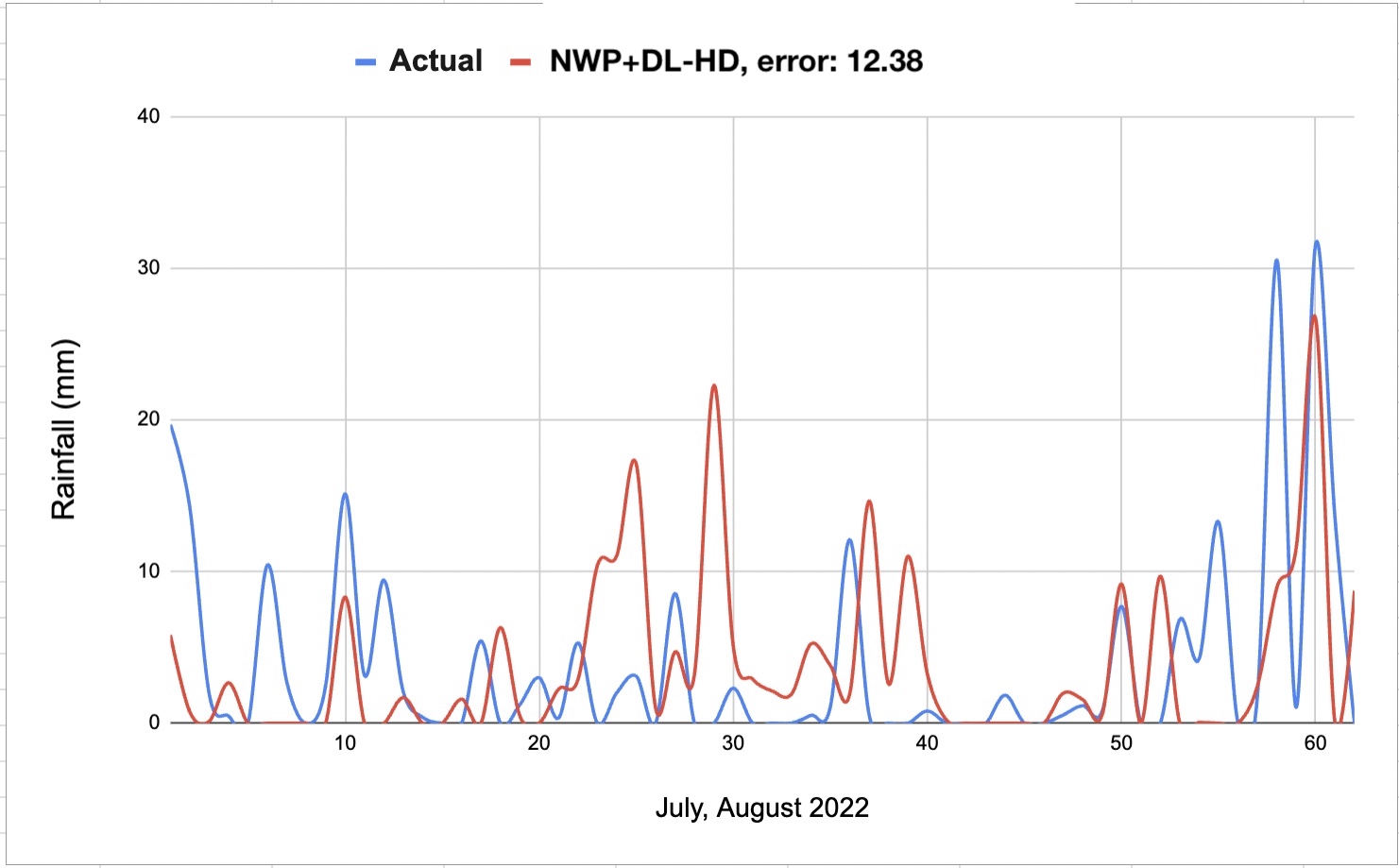}
    \caption{Ensemble vs IMD}

  \end{subfigure}

\end{figure}

\begin{figure}
  \caption{3-day forecasts for Mumbai in July and August 2022. None of the forecasts track the IMD ground truth well. However, DL-HD+Covariates capture some of the high and low rainfall events well, and we see a significant improvement using HRES-NWP+.}
  \label{fig:Mumbai_3day_loss}
  \centering
  \begin{subfigure}[b]{0.4\textwidth}
    \includegraphics[width=\textwidth]{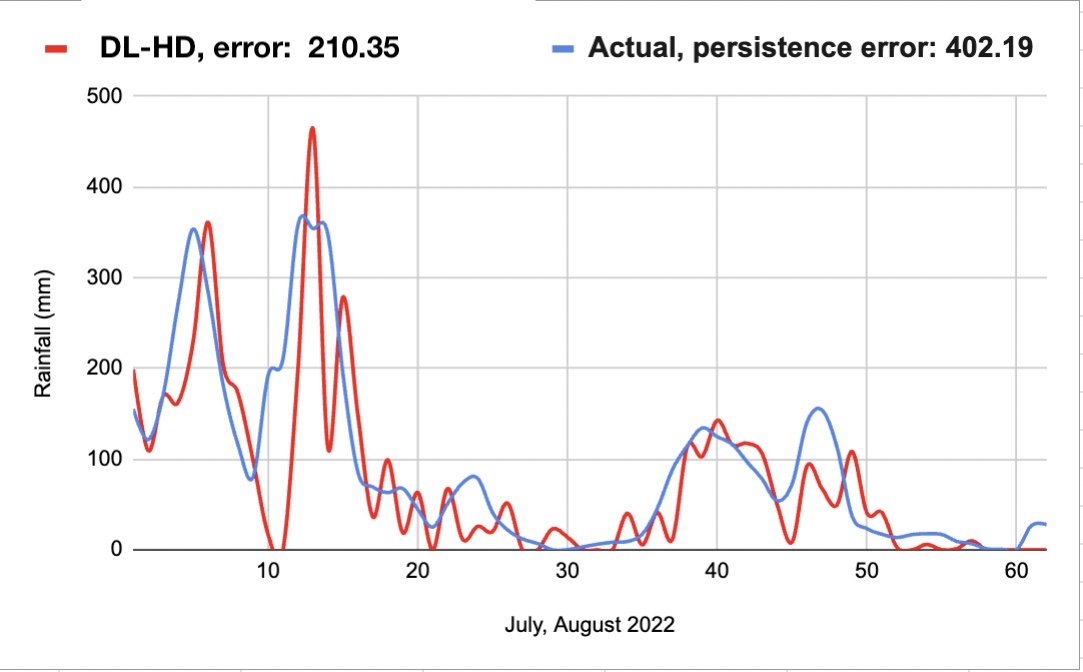}
    \caption{DL-HD+Covariates vs IMD}
    
  \end{subfigure}
  \begin{subfigure}[b]{0.4\textwidth}
    \includegraphics[width=\textwidth]{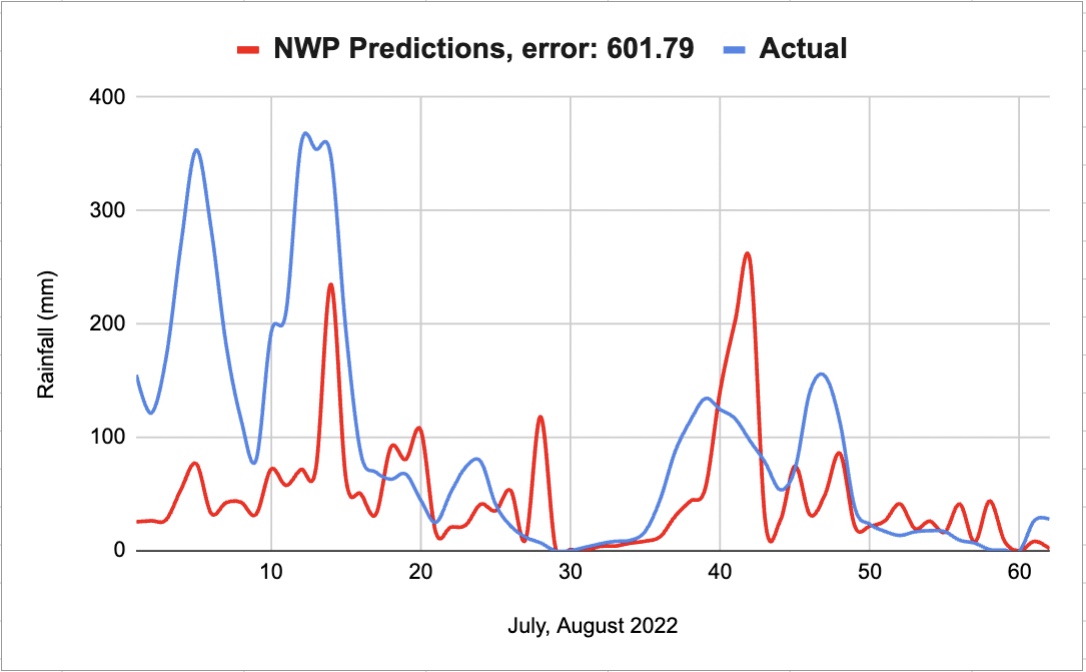}
    \caption{HRES vs IMD}
    
  \end{subfigure}
  
  \medskip
  
  \begin{subfigure}[b]{0.4\textwidth}
    \includegraphics[width=\textwidth]{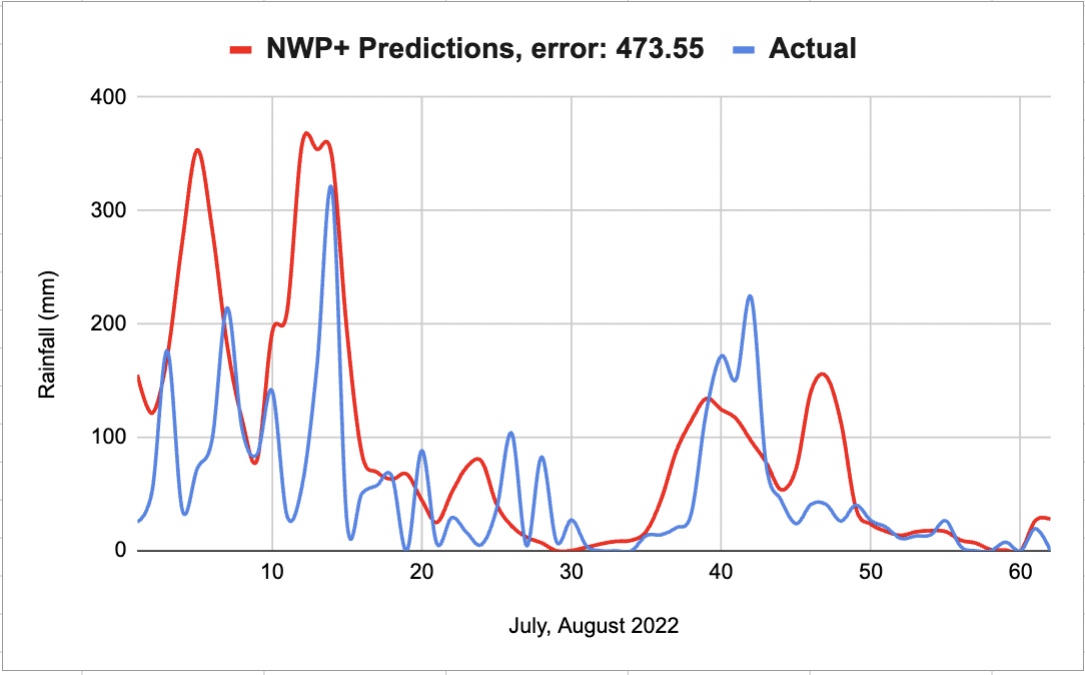}
    \caption{HRES-NWP+ vs IMD}
    
  \end{subfigure}
  \begin{subfigure}[b]{0.4\textwidth}
    \includegraphics[width=\textwidth]{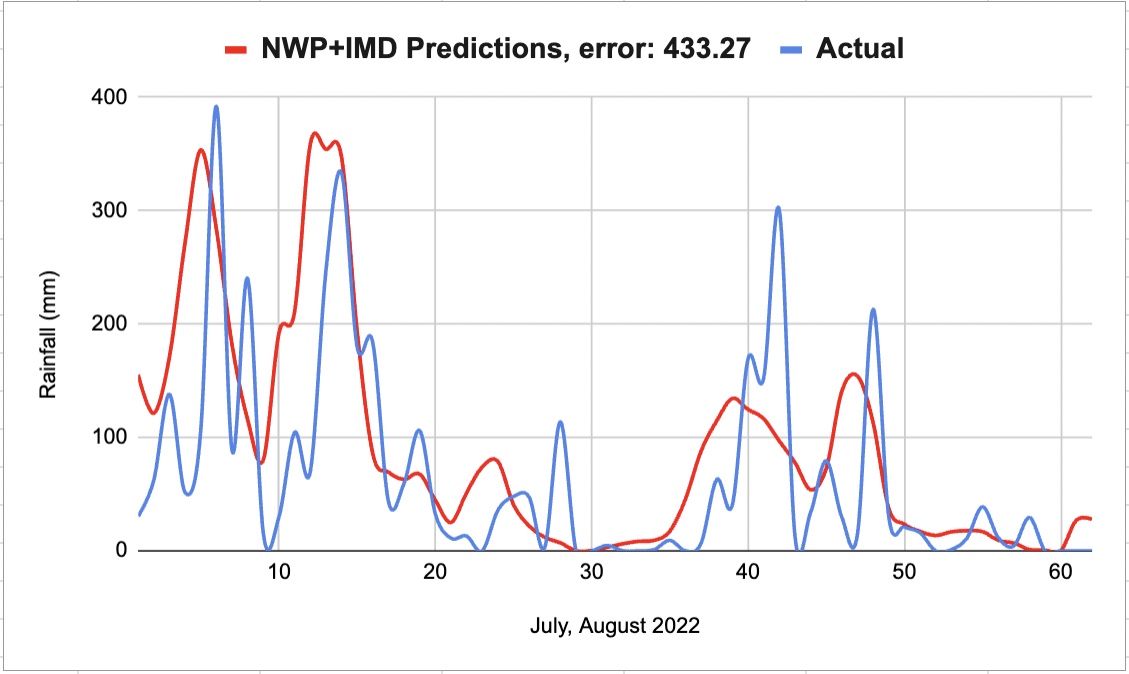}
    \caption{Ensemble vs IMD}
    
  \end{subfigure}

\end{figure}

\begin{figure}
  \caption{3-day forecasts for Ahmedabad in July and August 2022}
  \label{fig:Ahmedabad_3day_cities_loss}
  \centering
  \begin{subfigure}[b]{0.4\textwidth}
    \includegraphics[width=\textwidth]{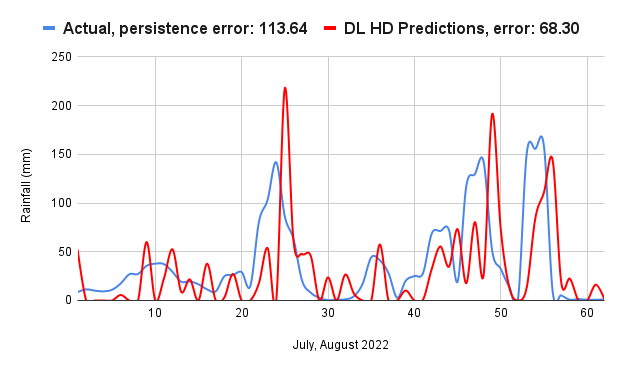}
    \caption{DL-HD+Covariates vs IMD}
  \end{subfigure}
  \begin{subfigure}[b]{0.4\textwidth}
    \includegraphics[width=\textwidth]{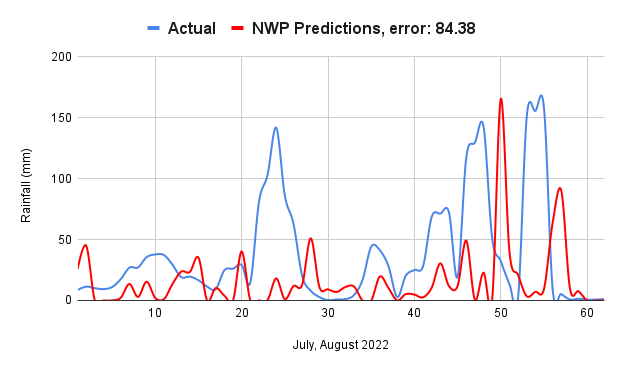}
    \caption{HRES vs IMD}
  \end{subfigure}
  
  \medskip
  
  \begin{subfigure}[b]{0.4\textwidth}
    \includegraphics[width=\textwidth]{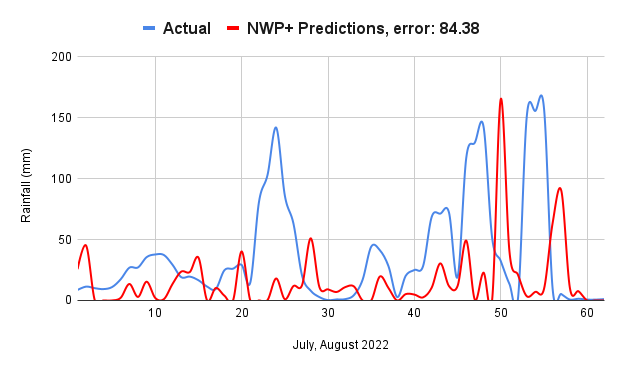}
    \caption{HRES-NWP+ vs IMD}
  \end{subfigure}
  \begin{subfigure}[b]{0.4\textwidth}
    \includegraphics[width=\textwidth]{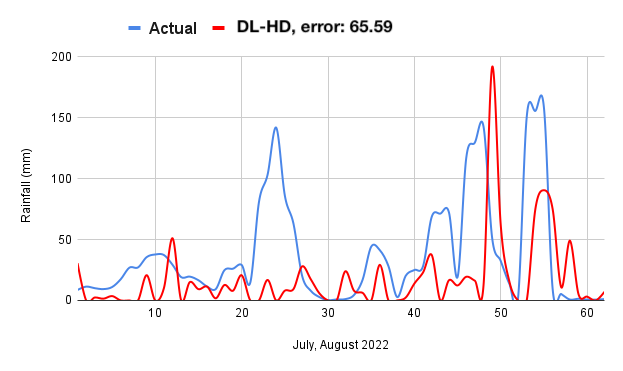}
    \caption{Ensemble vs IMD}
  \end{subfigure}

\end{figure}

\begin{figure}
  \caption{3-day forecasts for Chennai in July and August 2022}
  \label{fig:Chennai_3day_cities_loss}
  \centering
  \begin{subfigure}[b]{0.4\textwidth}
    \includegraphics[width=\textwidth]{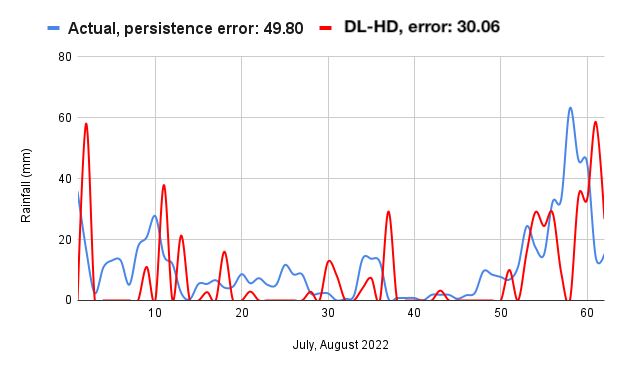}
    \caption{DL-HD+Covariates vs IMD}
  \end{subfigure}
  \begin{subfigure}[b]{0.4\textwidth}
    \includegraphics[width=\textwidth]{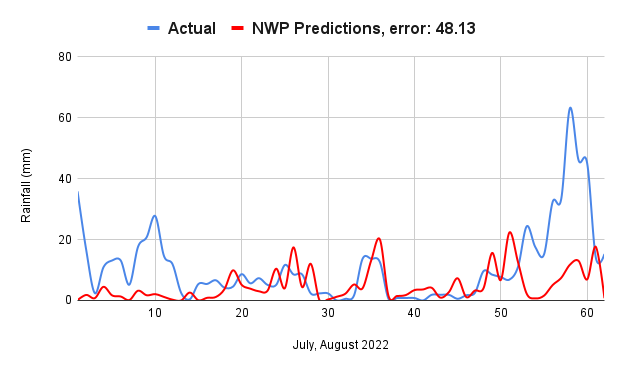}
    \caption{HRES vs IMD}
  \end{subfigure}
  
  \medskip
  
  \begin{subfigure}[b]{0.4\textwidth}
    \includegraphics[width=\textwidth]{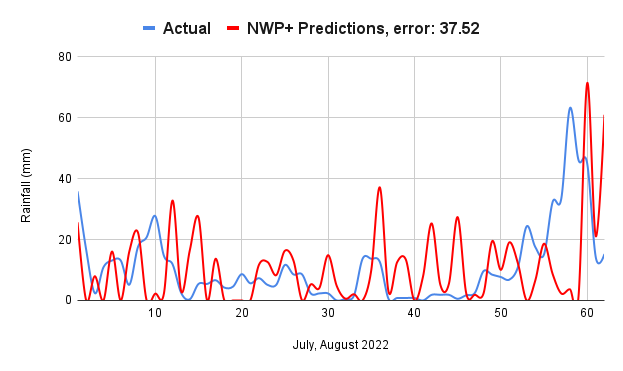}
    \caption{HRES-NWP+ vs IMD}
  \end{subfigure}
  \begin{subfigure}[b]{0.4\textwidth}
    \includegraphics[width=\textwidth]{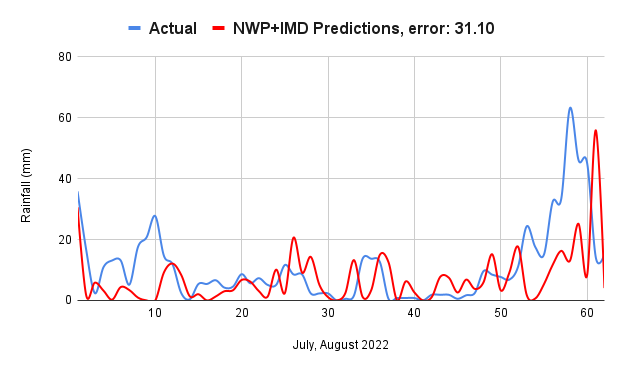}
    \caption{Ensemble vs IMD}
  \end{subfigure}

\end{figure}

Figures~\ref{fig:Mumbai_1day_cities_loss} to~\ref{fig:Chennai_1day_cities_loss} graphically compare the different forecasts with the ground truth for the cities Mumbai, Bhopal, Ahmedabad, and Chennai, for the months of July and August in 2022, for 1-day forecasts. Similar comparisons for 3-day forecasts are shown in Figures~\ref{fig:Mumbai_3day_loss} to~\ref{fig:Chennai_3day_cities_loss}. 

It is clear from the figures that DL-HD+Covariates forecasts consistently outperform other methods in tracking actual rainfall.

\begin{table}[h]
\caption{Comparison of 1-day ahead precipitation forecasting performance over India at 0.25° resolution.}
\label{tab:loss_india_1day_0.25deg}
\centering
\begin{tabular}{|l|c|c|c|}
    \hline
    \textbf{Model} & \textbf{Peak-biased Loss ($\mathrm{mm}^{1.5} + \mathrm{mm}$)} & \textbf{MSE ($\mathrm{mm}^2$)} & \textbf{\% Higher Error vs DL-HD + Covariates} \\
    \hline
    \textbf{DL-HD + Covariates} & 18.24 & 268.59 & - \\
    \textbf{DL-HD} & 20.90 & 312.11 & 16.21 \\
    \textbf{HRES-NWP} & 22.25 & 356.97 & 32.90 \\
    \textbf{HRES-NWP+} & 22.13 & 344.18 & 28.21 \\
    \textbf{Ensemble} & 18.96 & 294.25 & 9.56 \\
    \textbf{Persistence} & 25.42 & 448.10 & 39.36 \\
    \textbf{Climatological Mean} & 27.10 & 510.12 & 59.02 \\
    \textbf{Rolling Mean (20 Days)} & 29.75 & 563.20 & 81.68 \\
    \textbf{AR(1)} & 28.58 & 397.66 & 32.18 \\
    \textbf{AR(5) + Spatial Grids} & 27.22 & 364.21 & 30.98 \\
    \hline
\end{tabular}
\end{table}

\begin{table}[ht]
\caption{Comparison of 3-day ahead precipitation forecasting performance over India at 0.25° resolution.}
\label{tab:loss_india_3day_0.25}
\centering
\begin{tabular}{|l|c|c|c|}
    \hline
    \textbf{Model} & \textbf{Peak-biased Loss ($\mathrm{mm}^{1.5} + \mathrm{mm}$)} & \textbf{MSE ($\mathrm{mm}^2$)} & \textbf{\% Higher Error than DL-HD + Covariates} \\
    \hline
    \textbf{DL-HD + Covariates} & 67.28 & 2878.52 & - \\
    \textbf{DL-HD} & 81.87 & 3752.44 & 30.37 \\
    \textbf{HRES-NWP} & 85.59 & 4486.25 & 55.85 \\
    \textbf{HRES-NWP+} & 84.15 & 3884.24 & 34.89 \\
    \textbf{Ensemble} & 74.73 & 3019.81 & 4.91 \\
    \textbf{Persistence} & 114.06 & 8300.21 & 188.34 \\
    \textbf{Climatological Mean} & 120.50 & 8962.63 & 211.31 \\
    \textbf{Rolling Mean (20 Days)} & 126.75 & 9445.16 & 228.16 \\
    \textbf{AR(1)} & 106.42 & 5386.11 & 87.18 \\
    \textbf{AR(5) + Spatial Grids} & 83.45 & 4707.47 & 63.44 \\
    \hline
\end{tabular}
\end{table}

\newcommand{\minbold}[1]{\textbf{\boldmath#1}}

\begin{table}[ht]
\caption{Average peak-biased loss ($\text{mm}^{1.5}+\text{mm}$) for 1-day forecasts in grids corresponding to 20 major cities across India}
\label{tab:loss_1day_cities_0.25deg}
\centering
\begin{tabular}{|l|c|c|c|c|c|}
    \hline
    \textbf{City} & \textbf{DL-HD+Covariates} & \textbf{DL-HD} & \textbf{HRES-NWP} & \textbf{HRES-NWP+} & \textbf{Ensemble} \\
    \hline
    Ahmedabad & \textbf{16.23} & 18.46 & 25.34 & 23.15 & 17.32 \\
    Bangalore & 11.76 & 12.28 & 14.67 & 12.33 & \textbf{11.12} \\
    Bhopal & \textbf{24.35} & 26.76 & 28.92 & 27.14 & 23.98 \\
    Bhubaneswar & \textbf{26.42} & 29.11 & 28.04 & 27.91 & 27.74 \\
    Chandigarh & 18.45 & \textbf{18.35} & 20.42 & 20.18 & 19.65 \\
    Chennai & \textbf{10.33} & 11.15 & 12.40 & 12.26 & 12.18 \\
    Coimbatore & 10.27 & 10.77 & 11.55 & 11.31 & \textbf{10.09} \\
    Delhi & \textbf{7.44} & 7.86 & 12.79 & 11.77 & 8.85 \\
    Gangtok & \textbf{38.71} & 41.47 & 42.29 & 41.93 & 39.33 \\
    Hyderabad & \textbf{18.55} & 20.28 & 26.51 & 23.14 & 20.08 \\
    Indore & 11.22 & 11.41 & 15.65 & 15.41 & \textbf{10.40} \\
    Kochi & 19.26 & 22.77 & 26.13 & 25.42 & \textbf{19.07} \\
    Kolkata & \textbf{35.96} & 37.58 & 41.75 & 40.59 & 36.86 \\
    Lucknow & 11.86 & \textbf{11.72} & 16.46 & 16.11 & 11.99 \\
    Mumbai & \textbf{42.48} & 47.50 & 66.13 & 63.81 & 45.05 \\
    Patna & 12.71 & 13.33 & 15.58 & 15.51 & \textbf{12.46} \\
    Pune & \textbf{15.11} & 16.85 & 25.46 & 25.29 & 16.78 \\
    Raipur & \textbf{26.38} & 27.51 & 29.53 & 28.17 & 28.47 \\
    Shimla & \textbf{9.62} & 11.77 & 12.21 & 11.61 & 9.82 \\
    Vishakhapatnam & 23.49 & 26.75 & 27.21 & 27.18 & \textbf{22.53} \\
    \hline
    \textbf{Total Error} & \textbf{390.60} & 423.69 & 499.36 & 481.01 & 402.77 \\
    \textbf{\%age higher} & 0 & 8.38 & 27.77 & 23.05 & 3.08 \\
    \hline
\end{tabular}
\end{table}





\begin{table}[ht]
    \caption{Average peak-biased loss ($mm^{1.5}+ mm$) for 3-day forecasts in grids corresponding to 20 major cities across India}
    \label{tab:loss_3day_cities_0.25deg}
    \centering
    \begin{tabular}{|l|c|c|c|c|c|}
        \hline
        \textbf{City} & \textbf{DL-HD+Covariates} & \textbf{DL-HD} & \textbf{HRES-NWP} & \textbf{NWP+} & \textbf{Ensemble} \\
        \hline
        Ahmedabad & \textbf{58.89} & 65.22 & 66.29 & 65.96 & 59.88 \\
        Bangalore & 40.39 & 44.73 & 49.11 & 48.47 & \textbf{38.80} \\
        Bhopal & \textbf{72.25} & 79.16 & 82.43 & 81.57 & 75.49 \\
        Bhubaneswar & \textbf{56.76} & 69.46 & 74.03 & 74.14 & 64.32 \\
        Chandigarh & \textbf{45.87} & 50.25 & 55.27 & 53.34 & 48.19 \\
        Chennai & \textbf{33.53} & 43.10 & 51.54 & 49.78 & 40.22 \\
        Coimbatore & \textbf{42.15} & 47.36 & 50.96 & 50.48 & 46.14 \\
        Delhi & \textbf{32.16} & 32.53 & 40.53 & 39.81 & 32.11 \\
        Gangtok & \textbf{79.56} & 100.39 & 112.74 & 109.56 & 88.46 \\
        Hyderabad & \textbf{42.27} & 50.27 & 54.13 & 54.22 & 44.91 \\
        Indore & 57.42 & 57.29 & 62.94 & 60.41 & \textbf{56.71} \\
        Kochi & \textbf{59.49} & 69.56 & 74.21 & 74.44 & 62.24 \\
        Kolkata & \textbf{94.26} & 112.73 & 118.36 & 115.40 & 99.18 \\
        Lucknow & \textbf{30.71} & 34.58 & 53.91 & 51.76 & 32.15 \\
        Mumbai & \textbf{153.56} & 201.28 & 215.68 & 211.42 & 166.77 \\
        Patna & \textbf{29.93} & 34.22 & 41.39 & 41.14 & 30.28 \\
        Pune & \textbf{40.75} & 50.45 & 54.23 & 52.61 & 46.86 \\
        Raipur & \textbf{64.66} & 73.29 & 81.64 & 80.11 & 70.13 \\
        Shimla & \textbf{22.42} & 31.58 & 34.75 & 34.49 & 21.94 \\
        Vishakhapatnam & \textbf{72.48} & 80.56 & 76.50 & 75.93 & 76.54 \\
        \hline
        \textbf{Total Error} & \textbf{1130.85} & 1331.49 & 1449.77 & 1423.93 & 1223.65 \\
        \textbf{\%age higher} & 0 & 17.62 & 28.51 & 26.23 & 6.37 \\
        \hline
    \end{tabular}
\end{table}


\clearpage

\clearpage

\subsection{Additional performance comparisons}

In this section, we compare the performance of the DL-HD + Covariates model and HRES using confusion matrices (Figures \ref{tab:0perc-0mm}–\ref{tab:75perc-26mm}) computed across multiple rainfall thresholds, for the period 2022-2023. Specifically, we analyze confusion matrices in the 0th, 25th, 50th, and 75th rainfall percentiles to capture the behavior of the model over a wide range of rainfall intensities. These matrices provide detailed information on each model’s ability to correctly classify rainfall occurrences at varying thresholds.\\
To quantify classification performance, we report standard metrics derived from the confusion matrix: \textbf{Probability of Detection (POD)}, \textbf{False Alarm Ratio (FAR)}, \textbf{Probability of False Detection (POFD)}, and the \textbf{Critical Success Index (CSI)}. TP and FP denote true and false positives, respectively, and TN and FN denote true and false negatives. The POD measures the fraction of actual rainfall events that were correctly predicted as rain, and is computed as $\text{POD} = \frac{\text{TP}}{\text{TP} + \text{FN}}$, with higher values indicating better sensitivity to rainfall occurrences. The FAR quantifies the proportion of predicted rainfall events that did not actually occur, and is given by $\text{FAR} = \frac{\text{FP}}{\text{TP} + \text{FP}}$; a lower FAR implies improved precision by reducing the number of false alarms. The POFD captures the fraction of actual dry days that were incorrectly classified as rainy, calculated as $\text{POFD} = \frac{\text{FP}}{\text{FP} + \text{TN}}$, and is especially important for operational relevance, as a low POFD reduces unnecessary alerts. Finally, the CSI reflects the overall accuracy of rainfall predictions, penalizing both missed events and false alarms. It is defined as $\text{CSI} = \frac{\text{TP}}{\text{TP} + \text{FP} + \text{FN}}$, with higher values indicating more skillful and balanced classification performance.\\
In addition to classification skill, we also report the \textbf{Correlation Coefficient (CC)} between the predicted rainfall and the IMD ground truth across all grid points and time steps in Table \ref{tab:cc_1day}. It measures the linear relationship between the predicted and observed rainfall values, and is defined as: \textit{CC} $= \frac{\sum_i (P_i - \bar{P})(O_i - \bar{O})}{\sqrt{\sum_i (P_i - \bar{P})^2} \sqrt{\sum_i (O_i - \bar{O})^2}}$, where $P_i$ and $O_i$ denote the predicted and observed rainfall at index $i$, and $\bar{P}$ and $\bar{O}$ represent their respective means. CC values closer to 1 indicate stronger positive correlation, i.e., better agreement between the predicted and observed rainfall. 

Our results, again for the period 2022-2023 (Table \ref{tab:confusion_metrics}) demonstrate that DL-HD + Covariates consistently outperforms HRES at all examined rainfall percentile thresholds, both in confusion matrix statistics and derived skill scores. DL-HD+Covariates shows higher POD across all thresholds, reflecting better ability to detect rainfall events. The DL-HD+Covariates model also achieves lower FAR, indicating greater reliability in rain predictions. It also more effectively avoids false detection of rain during dry periods. Finally, higher CSI values demonstrate better overall classification performance when accounting for hits, misses, and false alarms.\\

\begin{figure}[ht]
\centering
\begin{minipage}[t]{0.48\textwidth}
\centering
\begin{tabular}{lcc}
\toprule
\textbf{DL-HD+Covariates} & \textbf{Actual $>$0 mm} & \textbf{Actual $\leq$0 mm} \\
\midrule
Predicted $>$0 mm & 136,394 (TP) & 15,154 (FP) \\
Predicted $\leq$0 mm & 15,154 (FN) & 1,348,776 (TN) \\
\bottomrule
\end{tabular}
\end{minipage}
\hfill
\begin{minipage}[t]{0.48\textwidth}
\centering
\begin{tabular}{lcc}
\toprule
\textbf{HRES} & \textbf{Actual $>$0 mm} & \textbf{Actual $\leq$0 mm} \\
\midrule
Predicted $>$0 mm & 128,500 (TP) & 18,200 (FP) \\
Predicted $\leq$0 mm & 22,100 (FN) & 1,355,000 (TN) \\
\bottomrule
\end{tabular}
\end{minipage}
\caption{Threshold = 0th Percentile (0 mm)}
\label{tab:0perc-0mm}
\end{figure}

\begin{figure}[ht]
\centering
\begin{minipage}[t]{0.48\textwidth}
\centering
\begin{tabular}{lcc}
\toprule
\textbf{DL-HD+Covariates} & \textbf{Actual $>$3.5 mm} & \textbf{Actual $\leq$3.5 mm} \\
\midrule
Predicted $>$3.5 mm & 21,170 (TP) & 6,109 (FP) \\
Predicted $\leq$3.5 mm & 6,109 (FN) & 1,475,096 (TN) \\
\bottomrule
\end{tabular}
\end{minipage}
\hfill
\begin{minipage}[t]{0.48\textwidth}
\centering
\begin{tabular}{lcc}
\toprule
\textbf{HRES} & \textbf{Actual $>$3.5 mm} & \textbf{Actual $\leq$3.5 mm} \\
\midrule
Predicted $>$3.5 mm & 19,800 (TP) & 7,400 (FP) \\
Predicted $\leq$3.5 mm & 8,200 (FN) & 1,480,500 (TN) \\
\bottomrule
\end{tabular}
\end{minipage}
\caption{Threshold = 25th Percentile (3.5 mm)}
\label{tab:25perc-3.5mm}
\end{figure}

\begin{figure}[ht]
\centering
\begin{minipage}[t]{0.48\textwidth}
\centering
\begin{tabular}{lcc}
\toprule
\textbf{DL-HD+Covariates} & \textbf{Actual $>$14 mm} & \textbf{Actual $\leq$14 mm} \\
\midrule
Predicted $>$14 mm & 12,500 (TP) & 3,200 (FP) \\
Predicted $\leq$14 mm & 4,800 (FN) & 1,512,000 (TN) \\
\bottomrule
\end{tabular}
\end{minipage}
\hfill
\begin{minipage}[t]{0.48\textwidth}
\centering
\begin{tabular}{lcc}
\toprule
\textbf{HRES} & \textbf{Actual $>$14 mm} & \textbf{Actual $\leq$14 mm} \\
\midrule
Predicted $>$14 mm & 11,300 (TP) & 4,100 (FP) \\
Predicted $\leq$14 mm & 5,600 (FN) & 1,518,000 (TN) \\
\bottomrule
\end{tabular}
\end{minipage}
\caption{Threshold = 50th Percentile (14 mm)}
\label{tab:50perc-14mm}
\end{figure}

\begin{figure}[ht]
\centering
\begin{minipage}[t]{0.48\textwidth}
\centering
\begin{tabular}{lcc}
\toprule
\textbf{DL-HD+Covariates} & \textbf{Actual $>$26 mm} & \textbf{Actual $\leq$26 mm} \\
\midrule
Predicted $>$26 mm & 8,400 (TP) & 1,900 (FP) \\
Predicted $\leq$26 mm & 2,300 (FN) & 1,530,000 (TN) \\
\bottomrule
\end{tabular}
\end{minipage}
\hfill
\begin{minipage}[t]{0.48\textwidth}
\centering
\begin{tabular}{lcc}
\toprule
\textbf{HRES} & \textbf{Actual $>$26 mm} & \textbf{Actual $\leq$26 mm} \\
\midrule
Predicted $>$26 mm & 7,800 (TP) & 2,500 (FP) \\
Predicted $\leq$26 mm & 3,100 (FN) & 1,538,000 (TN) \\
\bottomrule
\end{tabular}
\end{minipage}
\caption{Threshold = 75th Percentile (26 mm)}
\label{tab:75perc-26mm}
\end{figure}

\begin{table}[ht]
\caption{Comparison of classification metrics at multiple rainfall thresholds for DL-HD+Covariates and HRES.}
\label{tab:confusion_metrics}
\centering
\begin{tabular}{|c|l|c|c|c|c|}
\hline
\textbf{Threshold} & \textbf{Model} & \textbf{POD} & \textbf{FAR} & \textbf{POFD} & \textbf{CSI} \\
\hline
\multirow{2}{*}{0 mm} & DL-HD+Covariates & 0.900 & 0.100 & 0.011 & 0.818 \\
                      & HRES             & 0.853 & 0.124 & 0.013 & 0.761 \\
\hline
\multirow{2}{*}{3.5 mm} & DL-HD+Covariates & 0.776 & 0.224 & 0.004 & 0.634 \\
                        & HRES             & 0.707 & 0.272 & 0.005 & 0.559 \\
\hline
\multirow{2}{*}{14 mm} & DL-HD+Covariates & 0.723 & 0.204 & 0.002 & 0.610 \\
                       & HRES             & 0.669 & 0.266 & 0.003 & 0.538 \\
\hline
\multirow{2}{*}{26 mm} & DL-HD+Covariates & 0.785 & 0.184 & 0.001 & 0.667 \\
                       & HRES             & 0.716 & 0.243 & 0.002 & 0.582 \\
\hline
\end{tabular}
\end{table}

\begin{table}[ht]
\centering
\caption{Correlation coefficient (CC) of predicted rainfall with IMD ground truth for the period 2022–2023.}
\label{tab:cc_1day}
\begin{tabular}{l c}
\toprule
\textbf{Model} & \textbf{Correlation Coefficient (CC)} \\
\midrule
DL-HD + Covariates & \textbf{0.82} \\
DL-HD               & 0.75 \\
HRES           & 0.69 \\
NWP+                & 0.62 \\
Ensemble            & 0.81 \\
Persistence         & 0.49 \\
\bottomrule
\end{tabular}
\end{table}

\subsection{Spatio-temporal information in rainfall observations across India}
\label{subsec:lags}
\begin{figure}[ht]
    \caption{Comparison of average peak biased loss ($mm^{1.5}+ mm$) for coastal vs landlocked regions. In (a) the shaded region represents the grids spanning up to 60km from the coastline. (b) compares the error reduction with context for the different regions.}
    \label{fig:mainfig}
    \centering
    \begin{subfigure}[b]{0.43\textwidth}
        \centering
        \includegraphics[width=\textwidth]{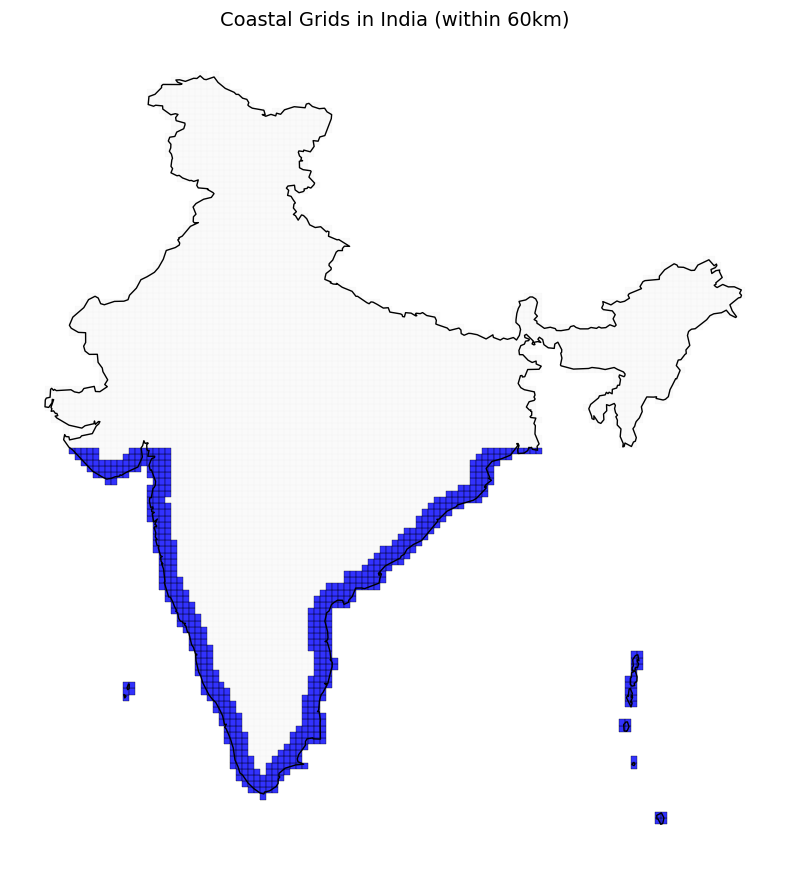}
        \caption{}
        \label{fig:subfig1}
    \end{subfigure}
    \hfill
    \begin{subfigure}[b]{0.47\textwidth}
        \centering
        \includegraphics[width=\textwidth]{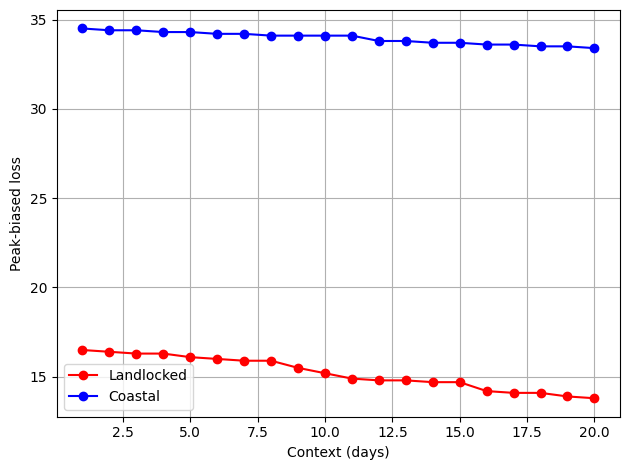}
        \caption{}
        \label{fig:subfig2}
    \end{subfigure}
\end{figure}

To examine the impact of historical context on forecast accuracy, we conducted experiments using input lags ranging from 3 to 20 days. We observe a pattern of small but consistently decreasing errors with increasing context length, indicating the presence of long-term memory in the data. To further analyze regional variation, we compare the aggregated performance across landlocked and coastal regions in Figure~\ref{fig:subfig2}. While increased historical context improves forecast accuracy in both regions, the gains are more pronounced in landlocked areas. This difference may be attributed to the availability of richer surrounding data in landlocked regions, whereas coastal areas are adjacent to oceanic regions where IMD precipitation data is unavailable. Incorporating oceanic rainfall data could potentially enhance forecast performance in coastal zones.


\subsection{Spatial performance and consistency}
\label{sec:spatial-consistency}
To assess spatial and annual consistency, we compute the number of grid points across India where the DL-HD+Covariates model produced lower daily forecast errors than the NWP baseline during the JJAS monsoon months for each year from 2017 to 2022. For each year, we calculate a \emph{win rate} by identifying, at each grid point, whether the DL-HD+Covariates model had a lower mean daily error compared to NWP. The total number of such grid points is then plotted annually.

\begin{figure}[ht]
    \centering
    \includegraphics[width=0.7\textwidth]{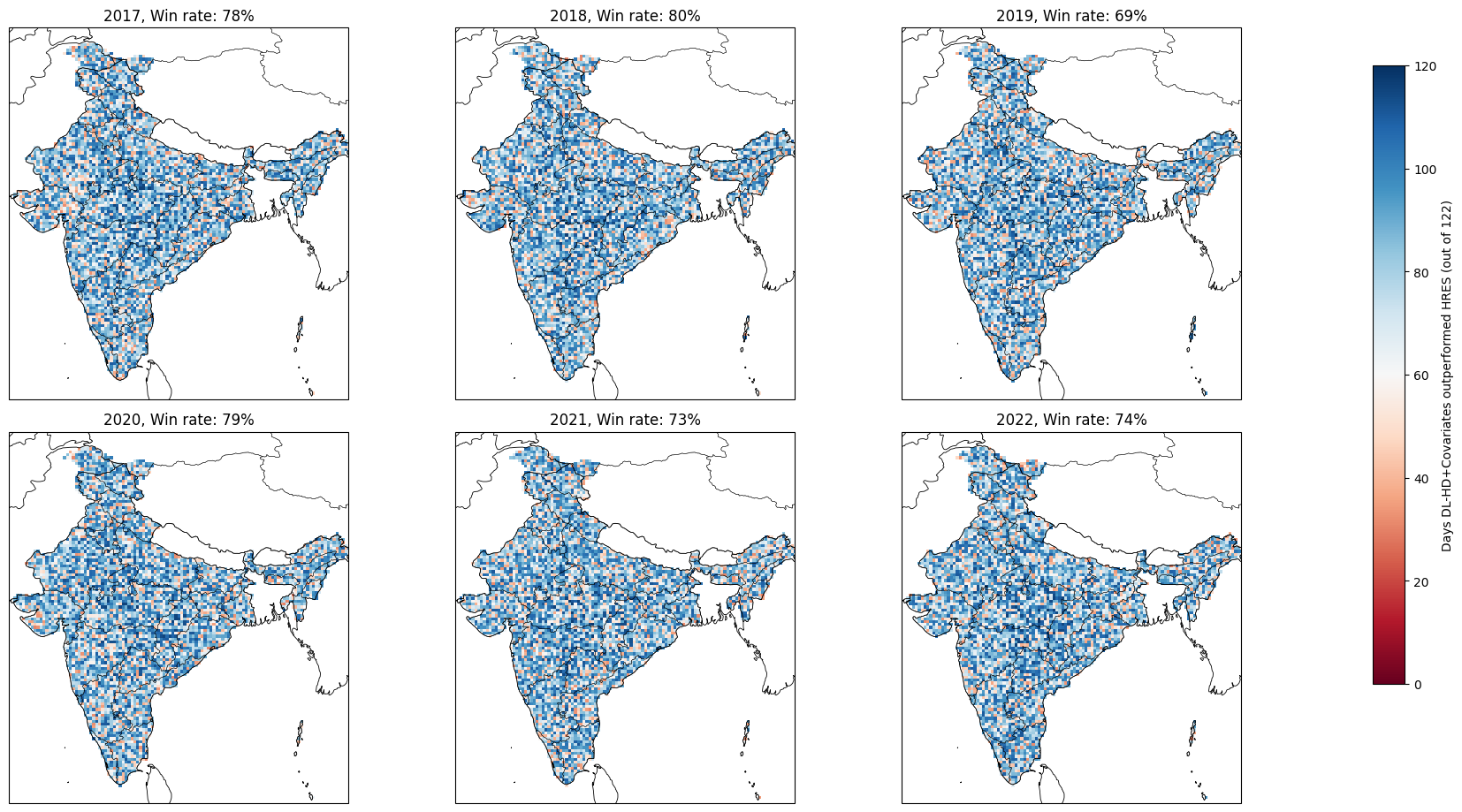}
    \caption{Annual count of grid points where the DL model produces lower daily forecast errors than NWP during JJAS from 2017 to 2022}
    \label{fig:spatial-consistency}
\end{figure}

Figure~\ref{fig:spatial-consistency} illustrates this win count for each monsoon season. We observe that the DL-HD+Covariates model wins in a majority of grid points across the country in every year,  demonstrating its spatial robustness and consistent outperformance of HRES-NWP.

\section{Discussion and Conclusion}

In this study, we demonstrated that deep learning models can consistently outperform traditional NWP systems, specifically the HRES and NCEP models, in forecasting monsoon rainfall over India, across spatial scales, lead times, and rainfall intensities. The DL-HD+Covariates model achieves a peak-biased loss of \textbf{18.24} and an MSE of \textbf{268.59}, surpassing HRES-NWP’s peak-biased loss of \textbf{22.25} (+22\% higher) and MSE of \textbf{356.97} (+32.9\% higher). For a lead time of 3 days, the gap widens: DL’s MSE of \textbf{2,878.52} contrasts sharply with HRES-NWP’s \textbf{4,486.25} (+55.85\% higher). These margins are consistent nationwide, and across key cities in India. The strong performance of deep learning (DL) models in rainfall forecasting can be explained by several physical factors. Traditional NWP models use physical equations to simulate the atmosphere. However, many small-scale processes, such as convection and cloud formation, occur at scales too fine to be directly resolved, so they are handled using simplified parameterizations. These approximations can introduce significant errors, especially over the Indian subcontinent, where weather patterns are highly complex and variable \citep{randall2003breaking, stensrud2007parameterization}. DL models, in contrast, do not rely on such physical approximations. Instead, they learn directly from historical data, identifying patterns that improve forecasts without needing to explicitly simulate physical processes. Another advantage of DL models is that they do not require precise initial conditions. NWP forecasts are highly sensitive to their initial inputs, and even small errors, particularly in regions with limited observations, such as oceans or mountainous areas, can grow quickly and reduce forecast accuracy \citep{lorenz1963deterministic, kalnay2003atmospheric}. DL models avoid this issue by using past sequences of observations to make predictions, making them more robust in data-sparse settings. Finally, the atmosphere behaves in a non-linear and sometimes chaotic way, which is difficult for traditional models to capture. DL architectures, especially those designed for spatiotemporal data are well-suited to handle this complexity. They can learn to represent these chaotic patterns, leading to more accurate and stable forecasts in challenging conditions.

\subsection{Possible future research directions}

Recent works \citep{kurz2024climode,zheng2025passat} have shown that incorporating physics into machine learning-based weather models can lead to models that are cheaper to train, more accurate, and faster to run. An exciting direction to extend the work of this paper is to develop such models for India using IMD data. 

A key direction suggested by our analysis is that more and
diverse data relevant to monsoon prediction, including radar- and satellite-based data \citep{espeholt2022deep}, combined
with carefully selected neural network architecture, are likely to substantially improve existing
NWP forecasts. However, radar-based data for public use are limited in India, and satellite data tend to be bulky and require substantially more computation. We leave incorporating radar-based data and satellite data into deep learning-based predictions using IMD data to a more expansive future project.

\clearpage
\acknowledgments
We would like to thank Sandip Trivedi, TIFR for motivating us to work on weather modelling and for related discussions.
We thank  Debasis Sengupta, ICTS,  Rama 
Govindrajan, ICTS and Partha Mukhopadhyay, IITM for
many helpful discussions. We also acknowledge the use of the High-Performance Computing (HPC) facility at TIFR for training large scale models required for this research.

%
%
\datastatement
The data utilized in this paper are publicly available and have been cited in Section \ref{sec:data}\ref{subsec: data_sources}.

\appendix

\subsection{Experiments for IMD data at $1^\circ \times 1^\circ$ resolution}
\label{app:1deg}
\subsubsection{Dataset Preparation}
While the  NCEP-NWP dataset shares the same spatial resolution as IMD at $1^{\circ} \times 1^{\circ}$, they are offset to each other by  $0.5^{\circ}$ in both latitude and longitude. IMD reports rainfall for grids whose latitudes and longitudes are multiples of $0.5^\circ$, for example the coordinates of a sample grid for IMD could be: [(79.5$^\circ$, 17.5$^\circ$), (80.5$^\circ$, 17.5$^\circ$), (79.5$^\circ$, 18.5$^\circ$), (80.5$^\circ$, 18.5$^\circ$)], while NCEP-NWP reports rainfall prediction for grids having integer coordinates. In order to find the best matching grid for IMD from NCEP-NWP, we first align the two by simply shifting all NWP grids by $0.5^{\circ}$ to the North-West. 
To establish a correspondence between each IMD grid and its NCEP-NWP counterpart, we identify the best-matching NCEP-NWP grid among the aligned set and its 4 surrounding grids, selecting the grid with the lowest forecast error.  

\subsubsection{Results}
\label{subsec: 1_degree}
We report the peak-biased loss corresponding to the different models for 1-day forecasts in Table \ref{tab:loss_india_1day_1deg}. Most trends are similar to those we observed for data at $0.25^\circ$ resolution, except that NCEP-NWP at $1^\circ \times 1^\circ$ consistently underperforms even persistence based forecasts. The errors for 3-day forecasts are reported in Table \ref{tab:loss_india_3day_1deg}. We also report the results for the key cities in Table \ref{tab:loss_1day_cities_1deg}.

\begin{table}[ht]
    \centering
    \begin{tabular}{|l|c|c|c|}
        \hline
        \textbf{Model} & \textbf{Peak-biased Loss} & \textbf{\%age Higher Error} \\
        \hline
        \textbf{DL-HD+Covariates} & 16.75 & - \\
        \textbf{DL-HD} & 18.11 & 8.12 \\
        \textbf{NCEP-NWP} & 23.98 & 43.16 \\
        \textbf{NCEP-NWP+} & 21.23 & 26.75 \\
        \textbf{Ensemble} & 16.88 & 0.78 \\
        \textbf{Persistence} & 23.05 & 37.61 \\
        \hline
    \end{tabular}
    \caption{Average peak-biased loss for 1-day forecasts across all the grids in India. NCEP-NWP exhibits the highest error, surpassing even the persistence model. The ensemble shows significant improvement over NWP alone, although it falls slightly short of DL-HD+Covariates.}
    \label{tab:loss_india_1day_1deg}
\end{table}

\begin{table}[ht]
    \centering
    \begin{tabular}{|l|c|c|c|}
        \hline
        \textbf{Model} & \textbf{Peak-biased Loss} & \textbf{\%age Higher Error} \\
        \hline
        \textbf{DL-HD+Covariates} & 67.02 & - \\
        \textbf{DL-HD} & 81.75 & 21.99 \\
        \textbf{NCEP-NWP} & 111.14 & 65.84 \\
        \textbf{NCEP-NWP+} & 104.46 & 55.85 \\
        \textbf{Ensemble} & 74.03 & 10.47 \\
        \textbf{Persistence} & 113.36 & 69.16 \\
        \hline
    \end{tabular}
    \caption{Average peak-biased loss for 3-day cumulative forecasts across all the grids in India. NWP exhibits the highest error, surpassing even the persistence model. Pooling NWP grids to form NWP+ leads to a substantial improvement in forecasts, yet it remains outperformed by DL-HD. The ensemble NWP+DL-HD shows significant improvement over NWP alone, although it falls slightly short of DL-HD.}
    \label{tab:loss_india_3day_1deg}
\end{table}

\begin{table}[ht]
    \centering
    \begin{tabular}{|l|c|c|c|c|c|}
        \hline
        \textbf{City} & \textbf{DL-HD+Covariates} & \textbf{DL-HD} & \textbf{NCEP-NWP} & \textbf{NCEP-NWP+} & \textbf{Ensemble} \\
        \hline
        Ahmedabad & \textbf{17.18} & 18.74 & 27.22 & 25.37 & 17.86 \\
        Bangalore & 12.19 & 13.55 & 17.56 & 15.11 & \textbf{12.14} \\
        Bhopal & \textbf{27.88} & 30.25 & 41.30 & 36.42 & 28.74 \\
        Bhubaneswar & \textbf{28.96} & 30.44 & 41.18 & 34.25 & 29.13 \\
        Chandigarh & 17.86 & \textbf{17.22} & 21.48 & 20.55 & 19.65 \\
        Chennai & \textbf{10.94} & 11.73 & 14.11 & 12.68 & 11.18 \\
        Coimbatore & 11.11 & 10.68 & 13.59 & 11.42 & \textbf{10.15} \\
        Delhi & \textbf{9.42} & 10.77 & 20.69 & 16.53 & 10.15 \\
        Gangtok & \textbf{36.44} & 39.04 & 42.88 & 40.54 & 38.87 \\
        Hyderabad & \textbf{21.03} & 22.28 & 31.12 & 27.58 & 22.06 \\
        Indore & 13.24 & 15.67 & 24.15 & 23.65 & \textbf{11.19} \\
        Kochi & 19.56 & 22.27 & 28.55 & 25.09 & \textbf{19.12} \\
        Kolkata & \textbf{35.14} & 37.63 & 41.78 & 38.82 & 36.66 \\
        Lucknow & 12.16 & \textbf{11.38} & 26.54 & 19.15 & 11.96 \\
        Mumbai & \textbf{44.42} & 47.53 & 86.88 & 71.89 & 45.05 \\
        Patna & 12.88 & 13.63 & 17.23 & 16.14 & \textbf{12.26} \\
        Pune & \textbf{15.11} & 16.85 &` 25.46 & 25.29 & 16.78 \\
        Raipur & \textbf{26.38} & 27.51 & 29.53 & 28.17 & 28.47 \\
        Shimla & \textbf{9.62} & 11.77 & 12.21 & 11.61 & 9.82 \\
        Vishakhapatnam & 23.49 & 26.75 & 27.21 & 27.18 & \textbf{22.53} \\
        \hline
        \textbf{Total Error} & \textbf{390.60} & 423.69 & 499.36 & 481.01 & 402.77 \\
        \textbf{\%age higher} & 0 & 8.38 & 27.77 & 23.05 & 3.08 \\
        \hline
    \end{tabular}

    \caption{Average peak biased loss for 1-day forecasts in grids corresponding to 20 major cities across India. Overall, and in most cities, DL-HD+Covariates outperforms other models by a significant margin. The ensemble model combining NWP and DL-HD+Covariates follows closely behind in the total error, and even performs better than just DL-HD+Covariates in some cities. HRES-NWP performs the poorest, having a $27.77\%$ higher error.}
    \label{tab:loss_1day_cities_1deg}
\end{table}






\subsection{Sensitivity Analysis}
To validate the design of our peak-biased loss function, we conducted a sensitivity analysis across different values of the parameters $\alpha$ and $\beta$. Tables~\ref{tab:confmat_0mm}-\ref{tab:confmat_26mm} report the model performance for different values of $\alpha$, and $\beta=1$, across 25-percentile rainfall thresholds.\\
The results indicate that $\alpha = 1.5$ and $\beta = 1.0$ provide the best trade-off. This analysis supports the empirical choice of asymmetry used in the main experiments and demonstrates the robustness of the proposed loss formulation.

\begin{table}[ht]
\centering
\caption{Confusion matrices for rainfall threshold $\geq 0$ mm (0th percentile) with different $\alpha$ values ($\beta=1$)}
\label{tab:confmat_0mm}
\begin{tabular}{|c|c|c|c|}
\hline
$\alpha=1.0, \beta=1$ & \textbf{Actual $\geq 0$ mm} & \textbf{Actual $= 0$ mm} \\
\hline
\textbf{Predicted $\geq 0$ mm} & 132,450 (TP) & 17,820 (FP) \\
\hline
\textbf{Predicted $= 0$ mm} & 19,098 (FN) & 1,341,110 (TN) \\
\hline
\end{tabular}

\begin{tabular}{|c|c|c|c|}
\hline
$\alpha=1.25, \beta=1$ & \textbf{Actual $\geq 0$ mm} & \textbf{Actual $= 0$ mm} \\
\hline
\textbf{Predicted $\geq 0$ mm} & 134,625 (TP) & 16,438 (FP) \\
\hline
\textbf{Predicted $= 0$ mm} & 16,923 (FN) & 1,342,492 (TN) \\
\hline
\end{tabular}

\begin{tabular}{|c|c|c|c|}
\hline
$\alpha=1.5, \beta=1$ & \textbf{Actual $\geq 0$ mm} & \textbf{Actual $= 0$ mm} \\
\hline
\textbf{Predicted $\geq 0$ mm} & 136,394 (TP) & 15,154 (FP) \\
\hline
\textbf{Predicted $= 0$ mm} & 15,154 (FN) & 1,343,776 (TN) \\
\hline
\end{tabular}

\begin{tabular}{|c|c|c|c|}
\hline
$\alpha=1.75, \beta=1$ & \textbf{Actual $\geq 0$ mm} & \textbf{Actual $= 0$ mm} \\
\hline
\textbf{Predicted $\geq 0$ mm} & 135,840 (TP) & 16,225 (FP) \\
\hline
\textbf{Predicted $= 0$ mm} & 15,708 (FN) & 1,342,705 (TN) \\
\hline
\end{tabular}
\end{table}

\begin{table}[ht]
\centering
\caption{Confusion matrices for rainfall threshold $\geq 3.5$ mm (25th percentile) with different $\alpha$ values ($\beta=1$)}
\label{tab:confmat_3.5mm}
\begin{tabular}{|c|c|c|c|}
\hline
$\alpha=1.0, \beta=1$ & \textbf{Actual $\geq 3.5$ mm} & \textbf{Actual $< 3.5$ mm} \\
\hline
\textbf{Predicted $\geq 3.5$ mm} & 19,875 (TP) & 7,230 (FP) \\
\hline
\textbf{Predicted $< 3.5$ mm} & 7,404 (FN) & 1,475,969 (TN) \\
\hline
\end{tabular}

\begin{tabular}{|c|c|c|c|}
\hline
$\alpha=1.25, \beta=1$ & \textbf{Actual $\geq 3.5$ mm} & \textbf{Actual $< 3.5$ mm} \\
\hline
\textbf{Predicted $\geq 3.5$ mm} & 20,635 (TP) & 6,750 (FP) \\
\hline
\textbf{Predicted $< 3.5$ mm} & 6,644 (FN) & 1,476,449 (TN) \\
\hline
\end{tabular}

\begin{tabular}{|c|c|c|c|}
\hline
$\alpha=1.5, \beta=1$ & \textbf{Actual $\geq 3.5$ mm} & \textbf{Actual $< 3.5$ mm} \\
\hline
\textbf{Predicted $\geq 3.5$ mm} & 21,170 (TP) & 6,109 (FP) \\
\hline
\textbf{Predicted $< 3.5$ mm} & 6,109 (FN) & 1,477,090 (TN) \\
\hline
\end{tabular}

\begin{tabular}{|c|c|c|c|}
\hline
$\alpha=1.75, \beta=1$ & \textbf{Actual $\geq 3.5$ mm} & \textbf{Actual $< 3.5$ mm} \\
\hline
\textbf{Predicted $\geq 3.5$ mm} & 20,843 (TP) & 6,875 (FP) \\
\hline
\textbf{Predicted $< 3.5$ mm} & 6,436 (FN) & 1,476,324 (TN) \\
\hline
\end{tabular}
\end{table}

\begin{table}[ht]
\centering
\caption{Confusion matrices for rainfall threshold $\geq 14$ mm (50th percentile) with different $\alpha$ values ($\beta=1$)}
\label{tab:confmat_14mm}
\begin{tabular}{|c|c|c|c|}
\hline
$\alpha=1.0, \beta=1$ & \textbf{Actual $\geq 14$ mm} & \textbf{Actual $< 14$ mm} \\
\hline
\textbf{Predicted $\geq 14$ mm} & 5,892 (TP) & 3,145 (FP) \\
\hline
\textbf{Predicted $< 14$ mm} & 3,201 (FN) & 1,498,240 (TN) \\
\hline
\end{tabular}

\begin{tabular}{|c|c|c|c|}
\hline
$\alpha=1.25, \beta=1$ & \textbf{Actual $\geq 14$ mm} & \textbf{Actual $< 14$ mm} \\
\hline
\textbf{Predicted $\geq 14$ mm} & 6,348 (TP) & 2,752 (FP) \\
\hline
\textbf{Predicted $< 14$ mm} & 2,745 (FN) & 1,498,633 (TN) \\
\hline
\end{tabular}

\begin{tabular}{|c|c|c|c|}
\hline
$\alpha=1.5, \beta=1$ & \textbf{Actual $\geq 14$ mm} & \textbf{Actual $< 14$ mm} \\
\hline
\textbf{Predicted $\geq 14$ mm} & 6,735 (TP) & 2,358 (FP) \\
\hline
\textbf{Predicted $< 14$ mm} & 2,358 (FN) & 1,499,027 (TN) \\
\hline
\end{tabular}

\begin{tabular}{|c|c|c|c|}
\hline
$\alpha=1.75, \beta=1$ & \textbf{Actual $\geq 14$ mm} & \textbf{Actual $< 14$ mm} \\
\hline
\textbf{Predicted $\geq 14$ mm} & 6,532 (TP) & 2,634 (FP) \\
\hline
\textbf{Predicted $< 14$ mm} & 2,561 (FN) & 1,498,751 (TN) \\
\hline
\end{tabular}
\end{table}

\begin{table}[ht]
\centering
\caption{Confusion matrices for rainfall threshold $\geq 26$ mm (75th percentile) with different $\alpha$ values ($\beta=1$)}
\label{tab:confmat_26mm}
\begin{tabular}{|c|c|c|c|}
\hline
$\alpha=1.0, \beta=1$ & \textbf{Actual $\geq 26$ mm} & \textbf{Actual $< 26$ mm} \\
\hline
\textbf{Predicted $\geq 26$ mm} & 3,106 (TP) & 1,254 (FP) \\
\hline
\textbf{Predicted $< 26$ mm} & 1,440 (FN) & 1,504,678 (TN) \\
\hline
\end{tabular}

\begin{tabular}{|c|c|c|c|}
\hline
$\alpha=1.25, \beta=1$ & \textbf{Actual $\geq 26$ mm} & \textbf{Actual $< 26$ mm} \\
\hline
\textbf{Predicted $\geq 26$ mm} & 3,384 (TP) & 1,097 (FP) \\
\hline
\textbf{Predicted $< 26$ mm} & 1,162 (FN) & 1,504,835 (TN) \\
\hline
\end{tabular}

\begin{tabular}{|c|c|c|c|}
\hline
$\alpha=1.5, \beta=1$ & \textbf{Actual $\geq 26$ mm} & \textbf{Actual $< 26$ mm} \\
\hline
\textbf{Predicted $\geq 26$ mm} & 3,728 (TP) & 818 (FP) \\
\hline
\textbf{Predicted $< 26$ mm} & 818 (FN) & 1,505,114 (TN) \\
\hline
\end{tabular}

\begin{tabular}{|c|c|c|c|}
\hline
$\alpha=1.75, \beta=1$ & \textbf{Actual $\geq 26$ mm} & \textbf{Actual $< 26$ mm} \\
\hline
\textbf{Predicted $\geq 26$ mm} & 3,525 (TP) & 936 (FP) \\
\hline
\textbf{Predicted $< 26$ mm} & 1,021 (FN) & 1,504,996 (TN) \\
\hline
\end{tabular}
\end{table}

\subsection{Neural Network Hyperparameters}
\label{app:nn_hyperparams}

This section describes the design and training setup of the two main models used in this study: a transformer-based model (Autoformer) and the simpler neural networks used for NWP+ and Ensemble models. Our choices were driven by strong empirical performance and computational efficiency.

\subsubsection{Autoformer Configuration}

We use the Autoformer model \citep{wu2021autoformer}, which is especially well-suited for making predictions over long time periods. It works by breaking down weather signals into different components and learning patterns over time using attention mechanisms. Key settings include:

\begin{itemize}
    \item \textbf{Transformer Layers:} Two layers in both the encoder and decoder, each using 8 attention heads. This allows the model to capture complex rainfall patterns across different time scales without becoming too heavy.
    
    \item \textbf{Embedding Size ($d_{\mathrm{model}}=512$):} This size balances detail and efficiency, for the 20-day input of 9 weather variables and helping the model learn interactions between them.
    
    \item \textbf{Feedforward Dimension ($d_{\mathrm{ff}}=2048$):} A larger internal layer helps the model learn complex relationships in atmospheric data.
    
    \item \textbf{Decomposition Kernel (Size 25):} This setting helps the model separate short-term fluctuations (like storms) from longer seasonal trends, which is important for understanding monsoon behavior.
\end{itemize}

\subsubsection{NWP+ Configuration}

NWP+ is a basic multilayer perceptron (MLP) that uses weather data from a central grid point, along with the 4 neighboring grid cells for better spatial context.

\begin{itemize}
    \item \textbf{Architecture:} A simple neural network with three layers, using 32, 16, and 1 neurons. This structure is compact because of limited data.
    
    \item \textbf{Activations:} ReLU is used in the hidden layers to handle spikes in rainfall, while the final output uses a sigmoid to keep predictions in a reasonable range after normalization.
    
    \item \textbf{Spatial Context:} Including nearby grid points helps improve the model’s accuracy, especially for short-term predictions. Adding more distant points didn’t help much, so we kept the neighborhood small.
\end{itemize}

\subsubsection{Training Strategy}

\begin{itemize}
    \item \textbf{Batch Size:} We use a larger batch (64) for the transformer to fully utilize GPU resources, and a smaller one (24) for the MLP due to memory limits.
    
    \item \textbf{Learning Rate:} The transformer uses a smooth cosine decay schedule, while the MLP uses a step-wise decrease every 50 epochs to help stabilize learning.
    
    \item \textbf{Regularization:} To avoid overfitting and training issues, we use weight decay and clip gradients that grow too large.
    
    \item \textbf{Early Stopping:} We monitor performance on a validation set and stop training if there's no improvement after 10–20 epochs. We also limit the training to a maximum of 100–300 epochs. 
    
    \item \textbf{Mixed Precision:} We train the Autoformer using half-precision (FP16), which speeds things up and reduces memory usage.
\end{itemize}

\subsubsection{Input and Output Processing}

\begin{itemize}
    \item \textbf{Normalization:} Input features are standardized using data from 2017–2021. This keeps the data consistent while still allowing year-to-year differences to be learned.

    \item \textbf{Prediction Target:} The models forecast daily rainfall for the next 1 to 3 days.
\end{itemize}


\bibliographystyle{ametsocV6}
\bibliography{references}

\end{document}